%% file: elsarticle-template-harv.tex
\documentclass[preprint,12pt,authoryear]{elsarticle}

\usepackage{graphicx}%
\usepackage{multirow}%
\usepackage{amsmath,amssymb,amsfonts}%
\usepackage{amsthm}%
\usepackage{mathrsfs}%
\usepackage[title]{appendix}%
\usepackage{xcolor}%
\usepackage{textcomp}%
\usepackage{manyfoot}%
\usepackage{booktabs}%
\usepackage{algorithm}%
\usepackage{algorithmicx}%
\usepackage{algpseudocode}%
\usepackage{listings}%
\usepackage{enumerate} 
\usepackage{color,soul}
\usepackage{subfigure}
\usepackage{changes}
\usepackage[normalem]{ulem}
\usepackage{url}
\usepackage{hyperref}
\usepackage{fixltx2e}
\usepackage{stfloats}
\usepackage{verbatim}
\usepackage{balance}

\journal{Nuclear Physics B}

\def \figref #1{\figurename \ \ref{#1}}
\def \tbref #1{\tablename \ \ref{#1}}
\def \eqref #1{Eq. (\ref{#1})}

\def \secref  #1{Section \ref{#1}}

\newcommand{\bm}[1]{\mbox{\boldmath $ #1 $}}
\newcommand{\pdif}[2]{\dfrac{\partial #1}{\partial #2}}

\begin{document}

\begin{frontmatter}



\title{Modeling and Control of a Rigid-soft Hybrid-link Humanoid Robot} 


\author{Zewen He\corref{cor1}\fnref{1}}\ead{he-zewen@ynl.t.u-tokyo.ac.jp}

\author{Taiki Ishigaki\fnref{1}}

\author{Ko Yamamoto\fnref{1}}

\cortext[cor1]{Corresponding author}

\affiliation[1]{organization={Department of Mechano-informatics, The University of Tokyo}, \addressline={7-3-1 Hongo}, city={Bunkyo-ku}, postcode={113-8656}, state={Tokyo}, country={Japan}}

\begin{abstract}
Physical flexibility plays a significant role not only in human motions but also in robot motions. 
For example, a humanoid robot with passive and flexible elements, which can be regarded as a rigid--soft hybrid-link system, has the advantages of storing energy and mitigating impact in a dynamic motion.
However, the motion design and control of a hybrid-link system are challenging to achieve and have only been discussed in the literature. 
In this study, we propose the framework of inverse kinematics and compliance optimization-based control for a hybrid-link humanoid system that represents the continuous deformation of a flexible structure using the piece-wise constant strain (PCS) model. 
The proposed method was validated using forward dynamics simulations for standing, jumping, and walking motions. 
Then, we built an experimental setup using a biped humanoid robot with leaf-spring type prosthesis leg, in which the prosthesis deformation was measured by a motion capture system.
Using this setup, we experimentally validated the proposed method for standing balance control in both double- and single-support cases. 
\end{abstract}




\end{frontmatter}

\input{introduction}
\input{fundamental}
\input{hik}
\input{hdcc}
\input{simulation}

\section{Conclusion \label{sect:conclusion}}

This study presented the fundamental framework of the control of hybrid-link humanoid system, including hybrid IK and compliance optimization control. 
The obtained results can be summarized as follows: 
\begin{enumerate}
    \item We proposed the hybrid IK that can explicitly consider passive elastic deformation due to the gravity. 
    In this method, an optimization problem that considers both elastic and gravity potential energy was designed. The proposed approach can find the optimal solution of the generalized coordinates satisfying the elastic-gravity force equilibrium using the Levenberg-Marquardt method. 
    \item We proposed a dynamically consistent compliance optimization framework for the hybrid-link system. 
    Both cases optimize the active compliance matrix with the given passive compliance. 
    In the forward dynamics simulation, the proposed dynamically consistent compliance optimization has been proven to have more advantages in balance control, and its control results have been further validated for jumping and walking motions. 
    In the jumping motion, the proposed method achieved 3 times higher jumping height than the reference trajectory without consideration of elasticity. 
    Through the comparison of joint torques between hybrid and rigid robot, we consider that the passive elasticity in the hybrid link system contributed to saving energy and enhancing locomotion capabilities. 
    In the walking motion. The error of COG in the $x$-direction was controlled within 0.16 m and the RMS error of PCS strain between hybrid IK and control results was 5.6\%. 
    \item We integrated the experimental platform involving a rigid-soft hybrid-link humanoid robot and a deformation measurement system based on motion capture equipment. 
    Based on this platform, we conducted a compliance control and the standing balance control in double-support phase. 
    During the double-support balance experiment, the COG displacement in $x$-direction was controlled within 0.5 cm. 
\end{enumerate}


\appendix
\input{appendix}










\bibliographystyle{elsarticle-harv.bst}
\bibliography{cas-refs}

\end{document}

%% file: introduction.tex
\section{Introduction}

\begin{figure}[t]
    \centering
\subfigure[]{
\centering
\includegraphics[width=0.18\hsize]{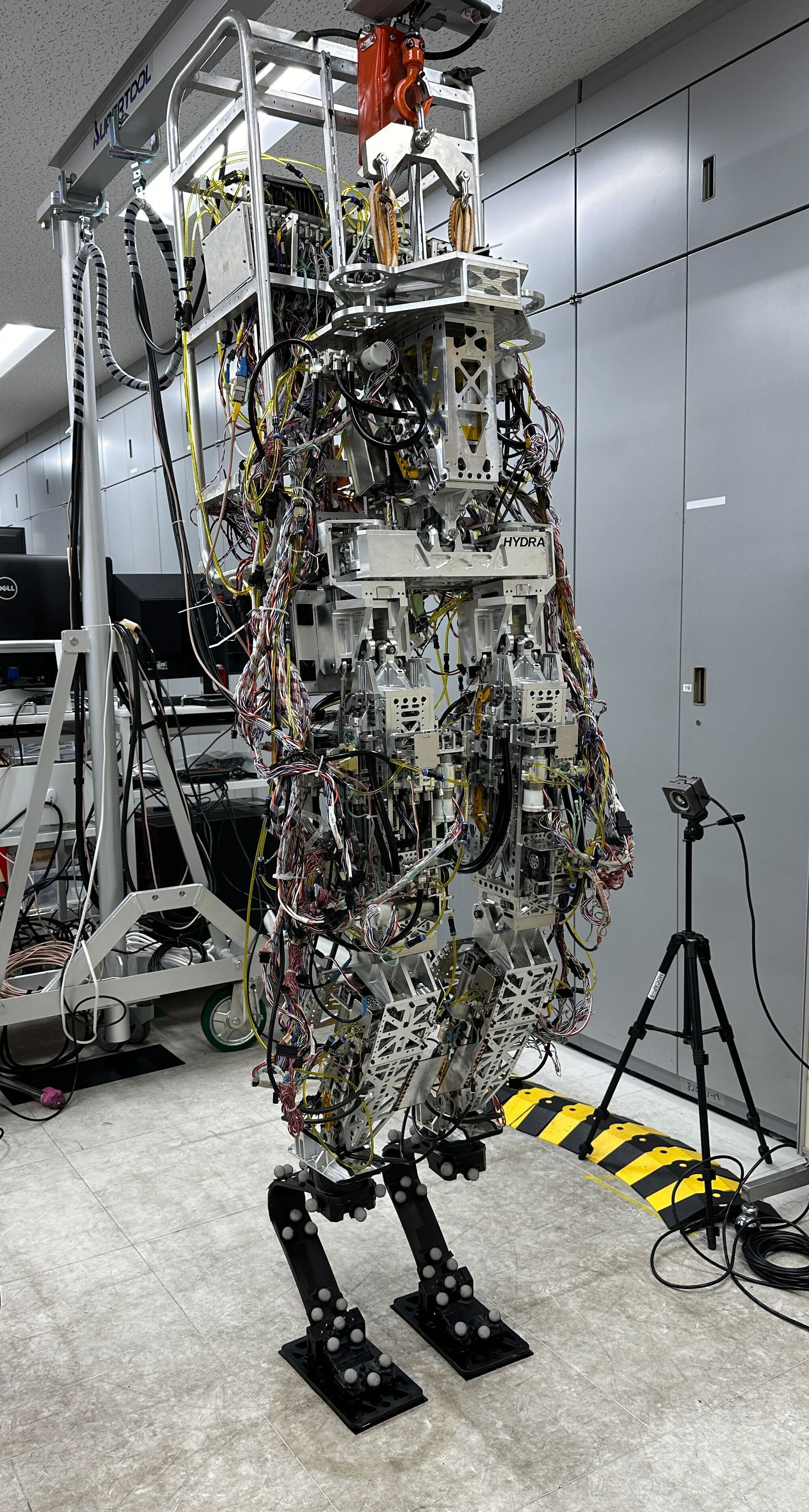}
\label{fig:hybrid_hydra}%
}\hspace{-2mm}
\subfigure[]{
\centering
\includegraphics[width=0.18\hsize]{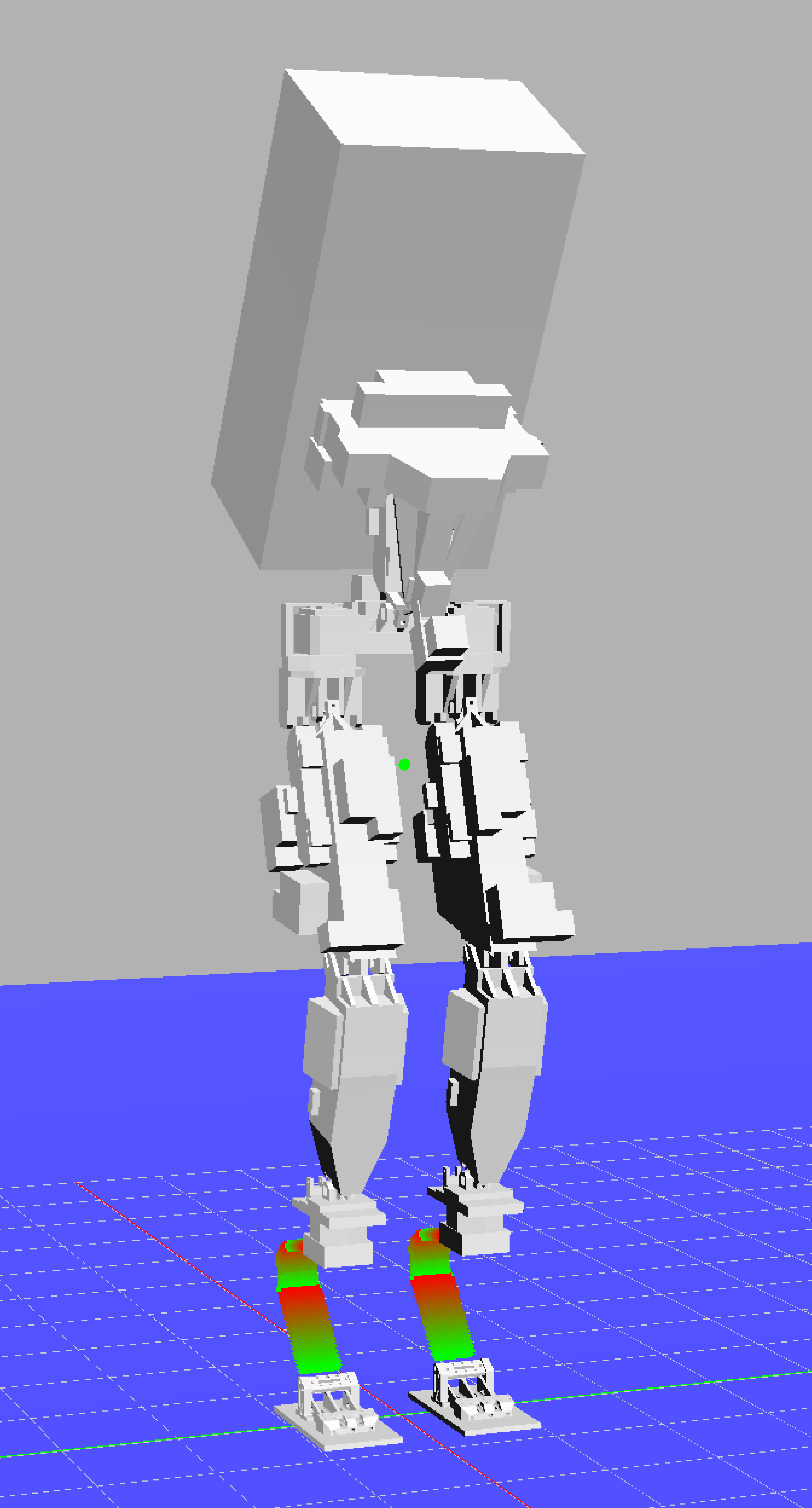}
\label{fig:hybrid_hydra_model}%
}\hspace{-2.0mm}
\subfigure[]{
\centering
\includegraphics[width=0.25\hsize]{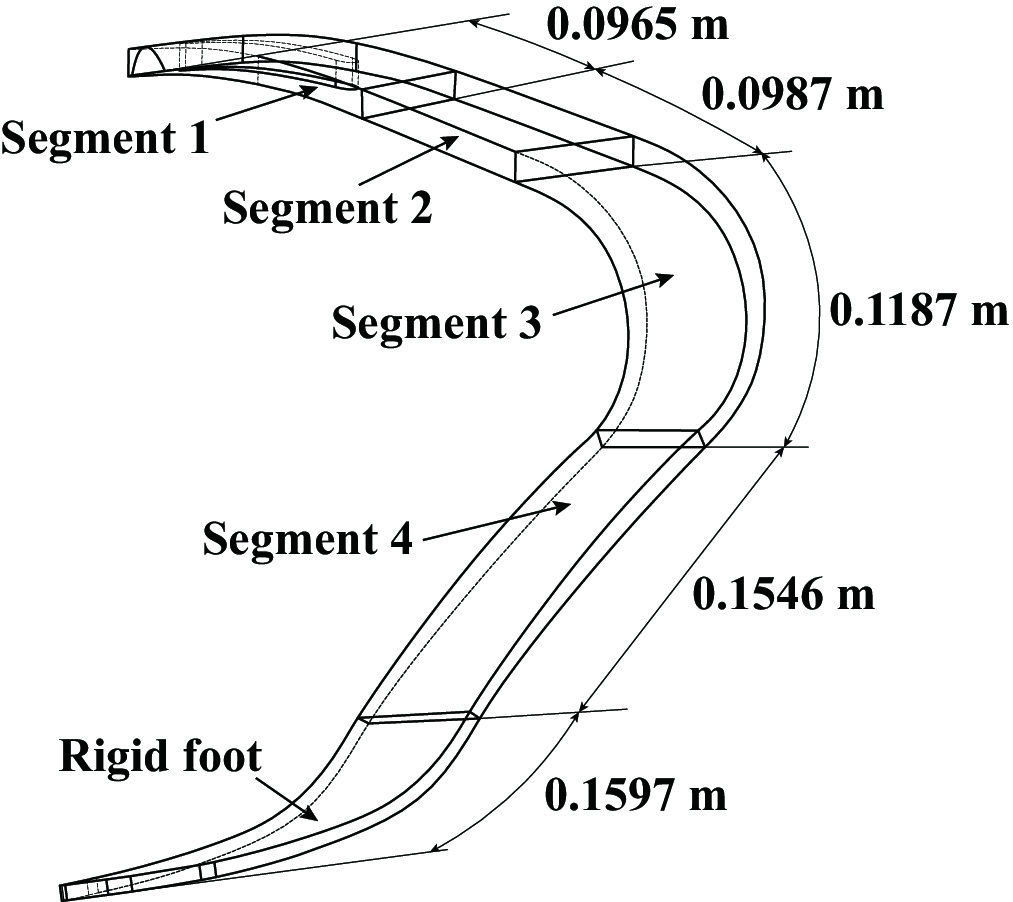}
\label{fig:1e91}%
}%

\subfigure[]{
\centering
\includegraphics[width=0.65\hsize]{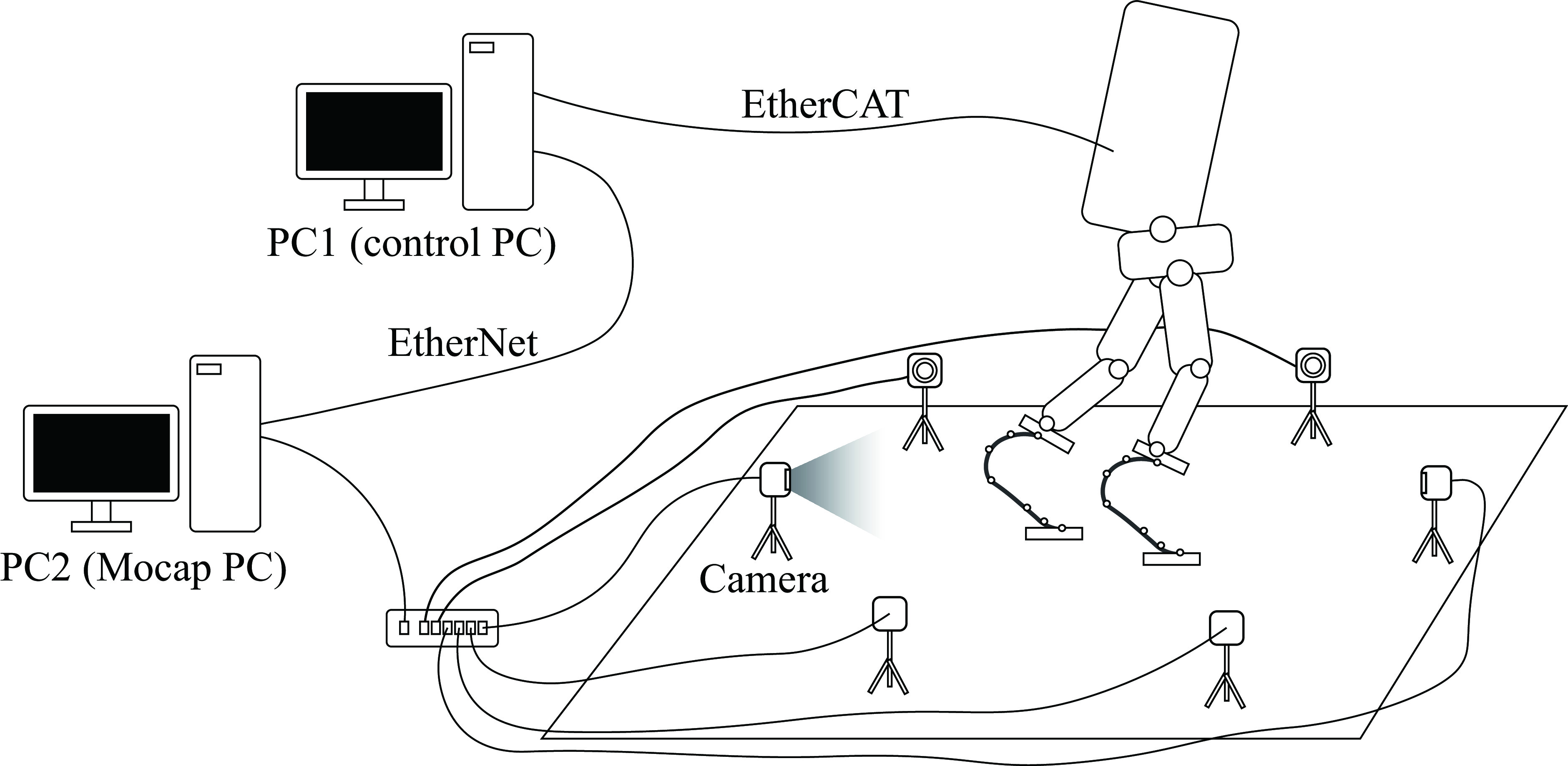}
\label{fig:hybrid_hydra_config}%
}
\caption{Overview of {\it hybrid Hydra}, which consists of a hydraulically-driven humanoid robot, Hydra \citet{ko2021whole}, wearing a leaf-spring type prosthesis at each leg: (a) its appearance, (b) hybrid-link model with PCS model of the prosthesis legs, (c) the detail of the PCS, and (d) experimental setup including motion capture system for measuring the deformation.}
\end{figure}

The human body system consists of rigid structures including skeleton and soft tissues including ligaments and tendons. 
In sports, the softness of specific shoes or flexible deformation of a leaf-spring type prosthesis contributes to impact absorption and energy storage during dynamic motion.
Stored energy can be used for acceleration, decreasing the required actuation force, which improves motion efficiency and dynamics \citet{vanderborght2013variable,laschowski2023energy}.
Although the control algorithms for most humanoid robots \citet{H1,Atlas} are based on the computation theory of a rigid-body multi-link system, numerous attempts have been made to utilize flexible elements in robotics \citet{pratt1995series,robinson1999series,werner2017optimal,niiyama2010athlete}.
Series elastic actuator (SEA) \citet{pratt1995series,robinson1999series} is commonly used in robot systems as a torque-controllable actuator, and its elasticity can be easily modeled by a linear spring \citet{werner2017optimal}.
On the other hand, humans can utilize a leaf-spring type prosthesis for dynamic motions despite its relatively larger and more complicated deformation.
One study attempted to apply this type of prosthesis to a humanoid robot \citet{niiyama2010athlete}.
However, because of the complexity of nonlinear dynamics of continuous deformation, the motion design and control of a humanoid robot with such passive flexibility remain challenging and have been rarely investigated, especially in the model-based approach.

Among the recent advances in soft robotics, discretized models based on the Cosserat theory \citet{Stuart2005Nonlinear} have been proposed for studying the deformation of a rod or beam structure \citet{Tunay2013TRO,renda2018discrete}. 
For example, the piece-wise constant strain (PCS) model \citet{renda2018discrete} discretizes a continuous deformation into several segments and assumes a constant strain at each segment. 
Furthermore, because the mathematical representation of PCS model is compatible with traditional rigid-body multi-link system, it was integrated into rigid-soft hybrid-link system  \citet{renda2018geometric,Ishigaki2021IROS}. 
This hybrid-link system can be used for modeling human motion wearing a leaf-spring type prosthesis \citet{kim2022inverse}.

However, motion control of such a hybrid-link system has been rarely investigated, especially for a humanoid robot equipped with a passive elastic element such as a flexible prosthesis.
In a humanoid system, a passive elastic element in leg undergoes a large deformation due to the gravity, which often leads to loss of body balance.
A pre-designed motion is required to maintain a stable balance despite this effect of gravity.
A similar problem must be solved to achieve real-time control.
An appropriate control input should be calculated for actuators considering the interaction between controlled joints and passive elements and the dynamic balance.

In this study, we integrate prosthetic legs into an electro-hydraulic humanoid robot to address joint power insufficiency and enhance energy efficiency. Therefore, the purpose of this study is to establish fundamental frameworks for motion control of a humanoid robot equipped with passive elasticity in its limbs, which can be modeled as hybrid-link systems.
The main contributions are summarized as follows: 
\begin{enumerate}
    \item Propose an inverse kinematics scheme for hybrid-link robot system to generate motion guaranteeing the equilibrium between the elastic restoring force and the gravity. 
    \item Propose a compliance optimization scheme for a hybrid-link humanoid robot considering passive viscoelasticity.
    \item Validate the feasibility of the proposed schemes and simultaneously substantiate the energy-saving capabilities of the elastic elements during dynamic motion, through simulations and experiments of a humanoid robot equipped with prostheses. 
\end{enumerate} 
One of the standard control methods for a humanoid system is the inverse dynamics control using quadratic programming method, often referred to as QP method \citet{Righetti2012Humanoids,bouyarmane2018quadratic}. 
Although this method ensures optimal distribution of contact forces the feasible solution of QP problem is very limited in a hybrid-link system with passive elastic elements.
This is further discussed in Section \ref{sect:qp_based_method_discussion}.
Rather, we directly deal with compliance matrices in the system so that active compliance is optimized, ensuring the total system stability including passive compliance.
Note that our control method is not implemented as position control but force/torque control because the optimized compliance matrices include non-zero value non-diagonal elements.
The proposed controller for standing balance, vertical hopping, and bipedal walking control was validated through forward dynamics simulation.
In the experiment, we constructed a hybrid-link humanoid robot platform and validated its balance control. 

The rest of this paper is organized as follows.
In \secref{sect:fundamental}, we introduce the hybrid-link system. 
Next, the proposed inverse kinematics method considering elastic deformation is illustrated in \secref{sec:hik}.
Then, the compliance optimization methods based on both statics and dynamics are proposed in \secref{sect:compliance}.  
\secref{sect:simulation} and \secref{sect:experiment} present the simulation and experiment results. 
The main simulation results have been published in \citet{he2024iros}. In this study, we added the walking motion simulation and experimentally validated its control. 
The conclusions are summarized in \secref{sect:conclusion}. 

As mentioned earlier, the first part of this paper has been published as a conference paper \citet{he2024iros}, which reported the basics of compliance-optimization control (\secref{sect:compliance}) and simulation results on the standing balance and hopping motion control (\secref{sect:simulation_double} and \ref{sect:simulation_jump}). 
The current study presents significant results on the inverse kinematics of the hybrid-link system, the simulation of biped walking motion control, and the experiment conducted to validate the simulation results through measurement of flexible deformations.

Hereafter, we use the following notations: 
\begin{itemize}
\item $\bm{O}$ and $\bm{E}$ denote the zero and unit matrices;
\item $\ast_{\rm ref}$ denotes the reference value of a quantity $\ast$;
\item $\varDelta \ast = \ast_{\rm ref} - \ast$ denotes the error between an arbitrary quantity $\ast$ and its reference value;
\item $\bm{X}^{\sharp}$ denotes the pseudo-inverse of $\bm{X}$. When $\bm{X}$ is horizontally-long and full row rank, $\bm{X}^{\sharp}$ can be calculated as $\bm{X}^{\sharp} = \bm{X}^T(\bm{X}\bm{X}^T)^{-1}$; when $\bm X$ is vertically-long and full column rank, $\bm{X}^{\sharp} = (\bm{X}^T \bm{X})^{-1}\bm{X}^T$; and
\item $\| \bm{x} \|^2_{\bm{W}} = \bm{x}^T \bm{W} \bm{x}$ is the weighted squared norm of a vector $\bm{x}$ with a weighting matrix $\bm{W}$.
\end{itemize}


%% file: fundamental.tex
\section{Rigid-soft Hybrid-link System \label{sect:fundamental}}

\subsection{Piece-wise Constant Strain Model \cite{renda2018discrete}}

\begin{figure}[t]
\centering
\subfigure[]{
\centering
\includegraphics[width=2.0in]{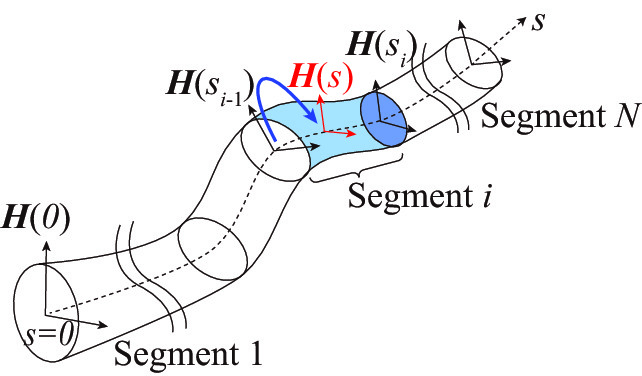}
\label{fig:pcs_model}
}\hspace{2mm}
\subfigure[]{
\centering
\includegraphics[width=1.8in]{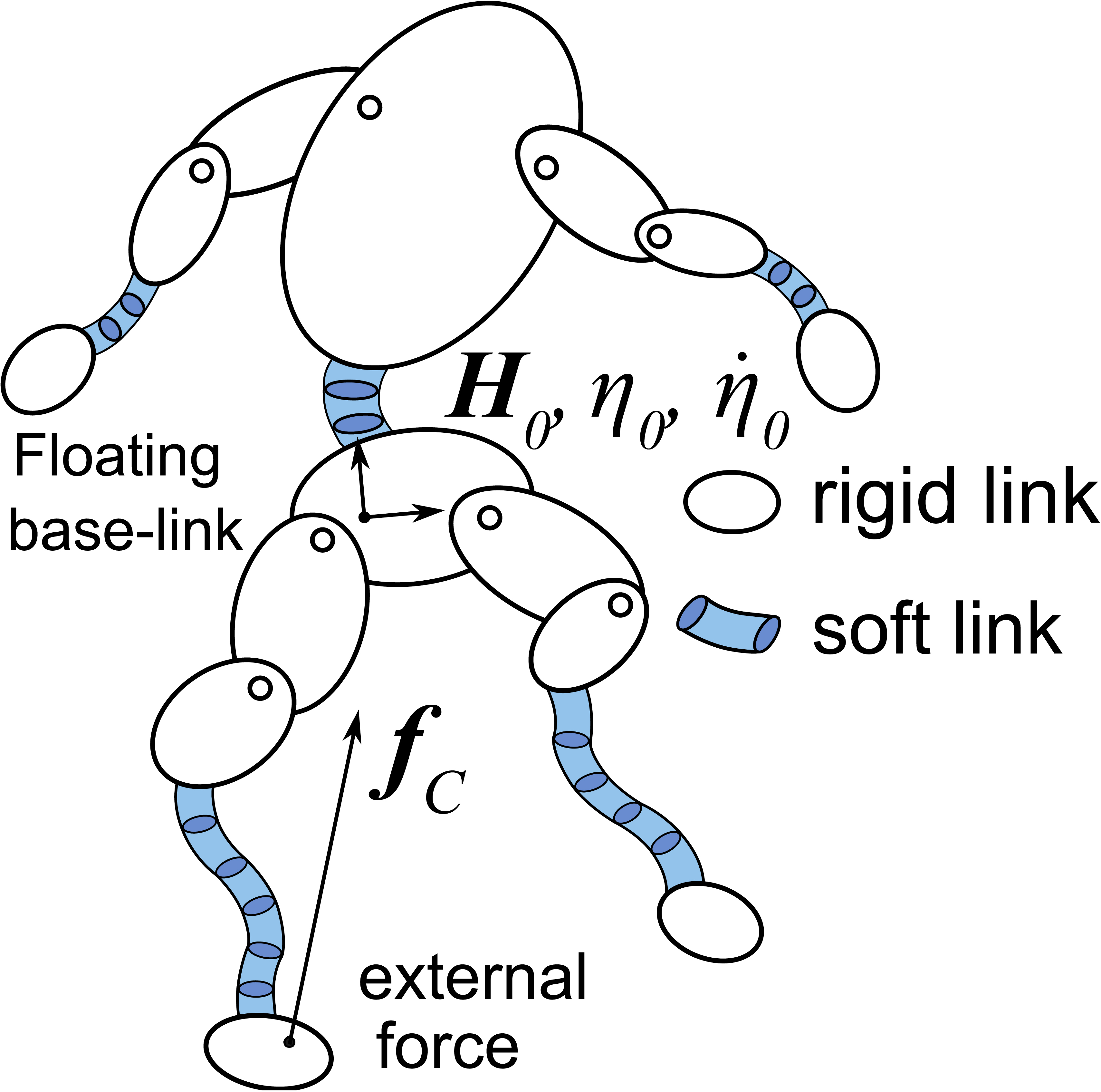}
\label{fig:hybrid_model}
}
\caption{(a) Schematic illustration of the PCS model \cite{renda2018discrete}. A beam or rod structure is divided into a finite number of {\it segments}. The six-dimensional strain is assumed to be spatially constant. (b) Illustration of a hybrid-link system. The blue parts represent the soft-part modeled using the PCS model, and the white parts represent rigid-part modeled using the rigid-body multi-link system. The base-link of the system is not fixed to the environment. An external force is considered to be acting on the end-effector of the system. }
\end{figure}

The PCS model discretizes a soft continuum into several constant-strain segments based on the Cosserat rod theory. 
In this study, we define the $z$-axis to be the direction of the central axis of the continuum. 
Let $s \in \mathbb R$ denote the central axis configuration, as shown in \figref{fig:pcs_model}. 
Then, we can express the configuration curve by defining a homogeneous transformation matrix $\bm{H}(s,t) \in {\rm SE(3)}$ at a certain time $t$ as follows: 
\begin{align}
    \bm{H}(s,t) = \begin{bmatrix} \bm{R}(s,t) & \bm p(s,t) 
    \\ \bm 0^T & 1 \end{bmatrix}
\end{align}
where, $\bm{R}(s,t) \in {\rm SO(3)}$ is orientation matrix; and 
$\bm p(s,t) \in \mathbb{R}^3$ is position vector. 
Hereafter, we omit $s$ or $t$ in the notation if its meaning is clear.

The deformation can be represented by a {\it strain} vector $\bm{\xi}(s) \in \mathbb{R}^6$, defined as an infinitesimal displacement in the configuration space that satisfies the following equation:
\begin{align}
    [ \bm{\xi}(s) \times ] = \bm{H}^{-1}(s) \pdif{\bm{H}(s)}{s}
    \in \mathfrak{se}(3)
\end{align}
The operator $[\ast \times ]$ is defined as
\begin{align}
    [\bm{\xi} \times ]
    :=
    \begin{bmatrix}
        [\bm{w} \times] & \bm{u} \\
        \bm{0}^T & 0
    \end{bmatrix},
    \quad
    \bm{\xi} 
    =
    \begin{bmatrix}
        \bm{w} \\ \bm{u}
    \end{bmatrix}
\end{align}
where, $\bm{w} \in \mathbb{R}^3$ and $\bm{u} \in \mathbb{R}^3$ are angular and linear strain vectors, respectively, and $[\bm{w} \times]$ is the skew-symmetric matrix for a 3D vector $\bm{w}$  that represents the outer product operation.
In the PCS model, the strain vector is assumed to be constant in each segment.
\begin{align}
    \bm{\xi}_i := \bm{\xi}(s) \ (s_i \le s \le s_{i+1})
\end{align}
where, $s_i$ denotes the center axis coordinate value of the starting point of the $i$-th segment with $s_0 = 0$.

Eventually, the generalized coordinate of the PCS model, $\bm{\theta}_S$, is defined as
\begin{align}
    \bm{\theta}_S :=
    \begin{bmatrix}
        (\bm{S}\bm{\xi}_1)^T & \cdots & (\bm{S}\bm{\xi}_{N_S})^T
    \end{bmatrix}^T \in \mathbb{R}^{n_S N_S}
\end{align}
where, $n_S$ is the number of DOFs of each segment, $N_S$ is the number of segments, and $\bm{S} \in \mathbb{R}^{n_S \times 6}$ is a selection matrix.
In the standard PCS model, we have $n_S = 6$ with $\bm{S} = \bm{E}$.
In the case of an inextensible Kirchhoff-Love segment, we have $n_S = 3$ with $\bm{S} = [\bm{E} \ \bm{O}]$, which means that the segment only has the angular strain $\bm{w}(s)$.

\subsection{Dynamics of a Hybrid-link System with Floating-base \cite{Ishigaki2021IROS}}

In \cite{Ishigaki2021IROS}, we proposed a hybrid-link system integrating a rigid-body multi-link system and flexible structure based on the PCS model, as shown in Fig. \ref{fig:hybrid_model}.
Hereafter, we refer to a part modeled by the PCS model as a {\it soft-part} and that modeled by the rigid-body multi-link system as a {\it rigid-part}.
Note that the soft-part can be located at any position in the hybrid-link system; it is not necessarily limited to the end.

Assuming that the base-link is not fixed to the environment, let $\bm{H}_0 \in {\rm SE(3)}$ and $\bm{\eta}_0 \in \mathbb{R}^6$ denote the homogeneous transformation matrix and spatial velocity vector of the base-link, respectively.
The generalized coordinates of the hybrid system and its velocity can be defined as
\begin{align}
    \bm{q} := \{ \bm{H}_0, \
    \begin{bmatrix}
    \bm{\theta}_R^T & \bm{\theta}_S^T        
    \end{bmatrix}
    ^T \}, \
    \dot{\bm{q}} :=
    \begin{bmatrix}
        \bm{\eta}_0^T & \dot{\bm{\theta}}_R^T & \dot{\bm{\theta}}_S^T
    \end{bmatrix}^T ,
\end{align}
where, $\bm{\theta}_R \in \mathbb R^{N_R}$ denote the generalized coordinates of the rigid-part and $N_R$ is the DOF of the rigid-part.
The total DOF of the system is $6 + N_R + n_SN_S$.

The dynamics of the hybrid-link system can be written as
\begin{align}
    \label{eq:floating_base_hybrid_system}
    \begin{bmatrix}
        \bm{M}_0 & \bm{M}_{0R} & \bm{M}_{0S}\\
        \bm{M}_{0R}^T & \bm{M}_R & \bm{M}_{RS} \\
        \bm{M}_{0S}^T & \bm{M}_{RS}^T & \bm{M}_S
    \end{bmatrix}
 \ddot{\bm{q}} +
    \begin{bmatrix}
    \bm{\beta}_0 \\ \bm{\beta}_R \\ \bm{\beta}_S 
    \end{bmatrix}
    = \begin{bmatrix}
       \bm{0} \\ \bm \tau_R \\ \bm \tau_S
    \end{bmatrix}
    + \mathcal{J}_C^T \bm{f}_C
\end{align}
where,
\begin{itemize}
    \item indexes 0, $R$, and $S$ denote that the corresponding quantity is associated with the base-link, rigid-part, and soft-part, respectively;
    \item $\bm{M}_{\ast}$ is a block element in the inertia matrix;
    \item $\bm{\beta}_{\ast}$ is a vector element in the bias term, which includes the centrifugal force, Coriolis force, and gravity;
    \item $\bm{\tau}_{\ast}$ is the generalized force of rigid- or soft-part;
    \item $\bm{f}_C$ is a vector that concatenates all contact force vectors $\bm{f}_{C,i} \in \mathbb{R}^6$ and is defined as
    \begin{align}
        \bm{f}_C :=
    \begin{bmatrix}
        \bm{f}_{C,1}^T & \cdots & \bm{f}_{C,N_C}^T
    \end{bmatrix}^T
    \in \mathbb{R}^{6N_C}
    \end{align}
    where $N_C$ is the number of contact points; and
    \item $\mathcal{J}_C^T \in \mathbb{R}^{(6 + N) \times 6N_C}$ ($N = N_R + n_S N_S$) is the Jacobian matrix corresponding to $\bm{f}_C$.
\end{itemize}

In this study, we assumed that the soft-part only had passive viscoelasticity without any actuation force. 
In this case, $\bm{\tau}_S$ is expressed as
\begin{align}
    \label{eq:passive_elasticity}
    \bm{\tau}_S = \bm{K}_S \varDelta \bm{\theta}_S - \bm{D}_S \dot{\bm{\theta}}_S,
    \\
    \varDelta \bm{\theta}_S := \bm{\theta}_{S,ref} - \bm{\theta}_S
\end{align}
where, $\bm{K}_S$, $\bm{D}_S \in \mathbb R^{n_S N_S\times n_S N_S}$ denote the stiffness and viscosity matrices, respectively, and are determined by material properties such as Young's modulus;
and $\bm{\theta}_{S, \rm ref} \in \mathbb R^{n_S N_S}$ represents the original coordinates of the soft part without deformation. 

The problem in controlling the hybrid system, as in \eqref{eq:floating_base_hybrid_system}, lies in generating the required control input $\bm{\tau}_R$ while considering the effects of passive elasticity expressed in \eqref{eq:passive_elasticity}, and the system stability.

%% file: hik.tex
\section{Inverse Kinematics for a Hybrid-link Humanoid Robot \label{sec:hik}}

\subsection{Hybrid IK that minimizes Potential Energies\label{sec:hik_single}} 
Motion generation of a hybrid-link humanoid system must account for the passive deformation due to gravity because such deformation often degrades the balance of a humanoid system. 
Accordingly, in this section, we present an inverse kinematics (IK) scheme for a hybrid-link system, referred to as {\it hybrid IK}. 

\begin{figure}[t]
\centering
\centering
\includegraphics[width=0.3\hsize]{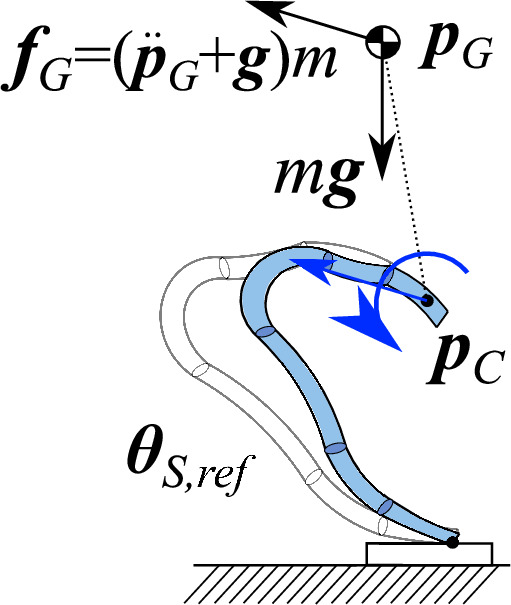}
\centering
\caption{Equilibrium between the gravity and elastic force in the hybrid-link system. The soft-part deforms due to the gravitational force, as indicated by the blue parts. \label{fig:hybrid_bal_2case-a}}
\end{figure}

We started the formulation of hybrid IK by assuming the single-support state of a hybrid-link humanoid, as shown in \figref{fig:hybrid_bal_2case-a}.
In this situation, a leg including a soft-part is in contact with the environment. 
The soft-part deforms due to the weight of the upper part.
The target of hybrid IK is to simultaneously calculate this passive deformation and the active joint coordinates when reference task-space positions are given.
In humanoid robot control, the task space is selected for the center of gravity (COG) or other limb positions.

In general, passive deformation is obtained as a result of minimizing the elastic energy and gravitational energy in the system. 
Therefore, the hybrid IK is formulated as an optimization problem as follows: 
\begin{align}
\label{eq:ikop}
 &\mathop{\rm min}\limits_{\bm q} g(\bm q) := \frac{1}{2} \lVert{\bm p_{\rm ref} - \bm p(\bm q)}\rVert_{\bm W_1}^2 + \frac{1}{2} \lVert{\Delta \bm \theta_S}\rVert_{\bm K_S}^2 + m \bm g^T \bm p_G \nonumber \\
 &\mathop{\rm s.t.} \quad \bm p_F(\bm q) - \bm p_{F,ref} = \bm 0  
\end{align}
where,
\begin{itemize}
    \item $\bm p(\bm{q})$ is a vector that concatenates position vectors of given multiple tasks including COG and foot positions; 
    \item $\bm p_F(\bm{q}) \in \mathbb R^{6}$ is the vector representing the position and orientation of the contact foot; 
    \item $m$ is the total mass of the robot;
    \item $\bm{g} = [ 0 \ 0 \ g ]^T$ is the gravitational acceleration vector; and
    \item $\bm p_G(\bm{q}) \in \mathbb R^3$ is the whole-body COG position
\end{itemize}
The second and third terms in the right-hand side of $g(\bm{q})$ represent elastic energy stored in the soft-part and the gravitational potential energy, respectively.

This is an optimization with an equality constraint and should be solved using the Lagrange method.
To avoid this, we assumed that the contact foot was fixed to the ground.
As a result, the vectors $\bm{p}(\bm{q})$ and $\bm{p}_G(\bm{q})$ become functions of $\bm{\theta} := [\bm{\theta}_R^T \ \bm{\theta}_S^T ]^T$.
Accordingly, the hybrid IK can be reformulated as
\begin{align}
\label{eq:ikop2}
    \min_{\bm{\theta}} \ g(\bm{\theta}) := 
    \frac{1}{2} \| \bm{p}_{ref} - \bm{p}(\bm{\theta}) \|_{\bm{W}}
    +
    \frac{1}{2} \| \varDelta \bm{\theta}_S \|_{\bm{K}_S}
    +
    m \bm{g}^T \bm{p}_G(\bm{\theta}) .
\end{align}
This optimization problem can solved using numerical methods such as Levenberg-Marquardt method \cite{sugihara2011solvability} without suffering from the singular configuration.

\subsection{Physical Interpretation of Gravity Term in Cost Function}
In this subsection, we selected the COG position as the task vector, namely $\bm{p} = \bm{p}_G$, to discuss the movement of the COG in the optimization process of \eqref{eq:ikop2}.
The gradient of the cost function $g(\bm{\theta})$ is calculated as\begin{align}
\label{grad2}
    \nabla g(\bm{\theta}) := \pdif{g}{\bm{\theta}}
    =
    -\varDelta \bm{p}_G^T \bm{W} \bm{J}_G(\bm{\theta})
    -\varDelta \bm{\theta}_S^T \bm{K}_S \bm{J}_S(\bm{\theta})
    +m\bm{g}^T \bm{J}_G(\bm{\theta})
\end{align}
where, $\bm{J}_S(\bm{\theta})$ and $\bm{J}_G(\bm{\theta})$ are Jacobian matrices defined as follows:
\begin{align}
    \bm{J}_G(\bm{\theta}) &:= \pdif{\bm{p}_G}{\bm{\theta}}
    =
    \begin{bmatrix}
        \bm{J}_{G,R} & \bm{J}_{G,S}
    \end{bmatrix}
    \\
    \bm{J}_S(\bm{\theta}) &:= \pdif{\bm{\theta}_S}{\bm{\theta}} 
    =
    \begin{bmatrix}
        \bm{O} & \bm{E}
    \end{bmatrix}
\end{align}
The matrices $\bm{J}_{\ast,R}$ and $\bm{J}_{\ast,S}$ denote block elements corresponding to $\bm{\theta}_R$ and $\bm{\theta}_S$, respectively.
If there exists an optimal solution $\bm{\theta} = \bm{\theta}_{\ast}$ that minimizes $g(\bm{\theta})$, the necessary condition $\nabla g^T(\bm{\theta}_\ast) = \bm{0}$ is represented as
\begin{align}
    \label{eq:necessary_condition01}
    \bm{J}_{G,R}^T (\bm{W} \varDelta \bm{p}_G - \bm{g}m) = \bm{0}
    \\
    \label{eq:necessary_condition02}
    \bm{J}_{G,S}^T(\bm{W} \varDelta \bm{p}_G  - \bm{g}m)+
    \bm{K}_S \varDelta \bm{\theta}_S = \bm{0}
\end{align}
\eqref{eq:necessary_condition02} implies that an equilibrium exists between the gravitational force and elastic force in the space of $\bm{\theta}_S$ if there exists an optimal solution that satisfies $\varDelta \bm{p}_G = \bm{0}$.
Note that this discussion can be extended to the case of the double-support state, as described in \ref{apdx:hybrid_ik_double}.

To further clarify the physical meaning of \eqref{eq:necessary_condition02}, we can divide $\bm{p}_G(\bm{\theta})$ as
\begin{align}
    \label{eq:cog_and_connecting_point}
    \bm{p}_G(\bm{\theta}) = \bm{p}_C(\bm{\theta}_S) + \bm{R}_C(\bm{\theta}_S) {}^{C}\bm{p}_{CG} (\bm{\theta}_R)
\end{align}
where, $\bm p_C(\bm{\theta}_S) \in \mathbb R^3$ is the position of the connecting point between rigid and soft parts;
$\bm R_C \in {\rm SO(3)}$ represents the orientation matrix of the local frame $\{C\}$ fixed on $\bm p_C$; and $\sideset{^C}{_{CG}}{\mathop{\bm{p}}}$ is the relative position vector from $\bm p_C$ to $\bm p_G$ w.r.t. $\{C\}$. 
Note that $\bm{p}_C(\bm{\theta}_S)$ and $\bm{R}_C(\bm{\theta}_S)$ depend only on $\bm{\theta}_S$, and ${}^{C}\bm{p}_{CG}(\bm{\theta}_R)$ depends only on $\bm{\theta}_R$.
\figref{fig:foot_coord} illustrates the coordinates of the connecting link. 

From \eqref{eq:cog_and_connecting_point},
\begin{align}
    \label{eq:necessary_condition03}
    & \bm{J}_{G,R}^T\bm{W} \varDelta \bm{p}_G - \bm{J}_{CG,R}^T \bm{R}_C^T \bm{g}m = \bm{0}
    \\
    \label{eq:necessary_condition04}
    & \bm{J}_{G,S}^T\bm{W} \varDelta \bm{p}_G + \bm{K}_S \varDelta \bm{\theta}_S - \bm{J}_{C,S}^T \bm{g}m - \bm{J}_{\omega,S}^T \bm{p}_{CG} \times \bm{g}m = \bm{0}
\end{align}
Derivation of these equations is described in \ref{apdx:necessary_condition_derivation}.

\eqref{eq:necessary_condition04} represents an equilibrium among the elastic force, the gravity force, and the moment by the gravity force if there exists an optimal solution that satisfies $\varDelta \bm{p}_G = \bm{0}$.
If there is a residual error of $\varDelta \bm{p}_G$, the weighting matrix $\bm{W}$ can be regarded as a {\it virtual} stiffness matrix for the COG position, and the first term in \eqref{eq:necessary_condition04} represents a restoring force to $\bm{p}_{G,ref}$.
Similarly, \eqref{eq:necessary_condition03} represents an equilibrium between the gravity force and the restoring force resulting from virtual stiffness.

Setting a larger value for $\bm{W}$, ideally infinite, is desirable to make $\Delta {\bm p}_G$ zero. 
In practice, however, such a large value usually makes the optimization unstable.
Whereas, a smaller value of $\bm{W}$ that allows some residual error stabilizes the numerical calculation. 
However, the value of $\bm{W}$ is smaller than that of $\bm{K}_S$, and the error $\varDelta \bm{p}_G$ increases significantly.
For example, we observed this error to grow so large that the COG almost reached the ground.

To avoid this problem, we set ${}^C \bm{p}_{CG}$ to a constant vector during the optimization process.
This makes the necessary condition \eqref{eq:necessary_condition03} satisfied even if the residual error $\varDelta \bm{p}_G$ is not zero because $\bm{J}_R = \bm{J}_{CG,R} = \bm{O}$.

\subsection{Hybrid IK for Bipedal Walking \label{sect:hik_acc}}
In the previous subsection, we assumed a static equilibrium in the hybrid IK.
In humanoid robot control, a trajectory of the COG can be generated based on the simplified COG-ZMP model \cite{sugihara20213d}, which includes the COG acceleration for a dynamic walking or hopping motion. 
We can consider such a dynamical effect by replacing $m\bm{g}$ in \eqref{eq:ikop} with
\begin{align}
 \bm f_G := (\ddot{\bm p}_{G, \rm ref} + \bm g) m
\end{align}
where $\ddot{\bm p}_{G, \rm ref} \in \mathbb R^3$ is the reference value of the COG acceleration. 

\begin{figure}[t]
    \begin{center}
    \subfigure[]{
    \centering
	 \includegraphics[width=0.22\hsize]{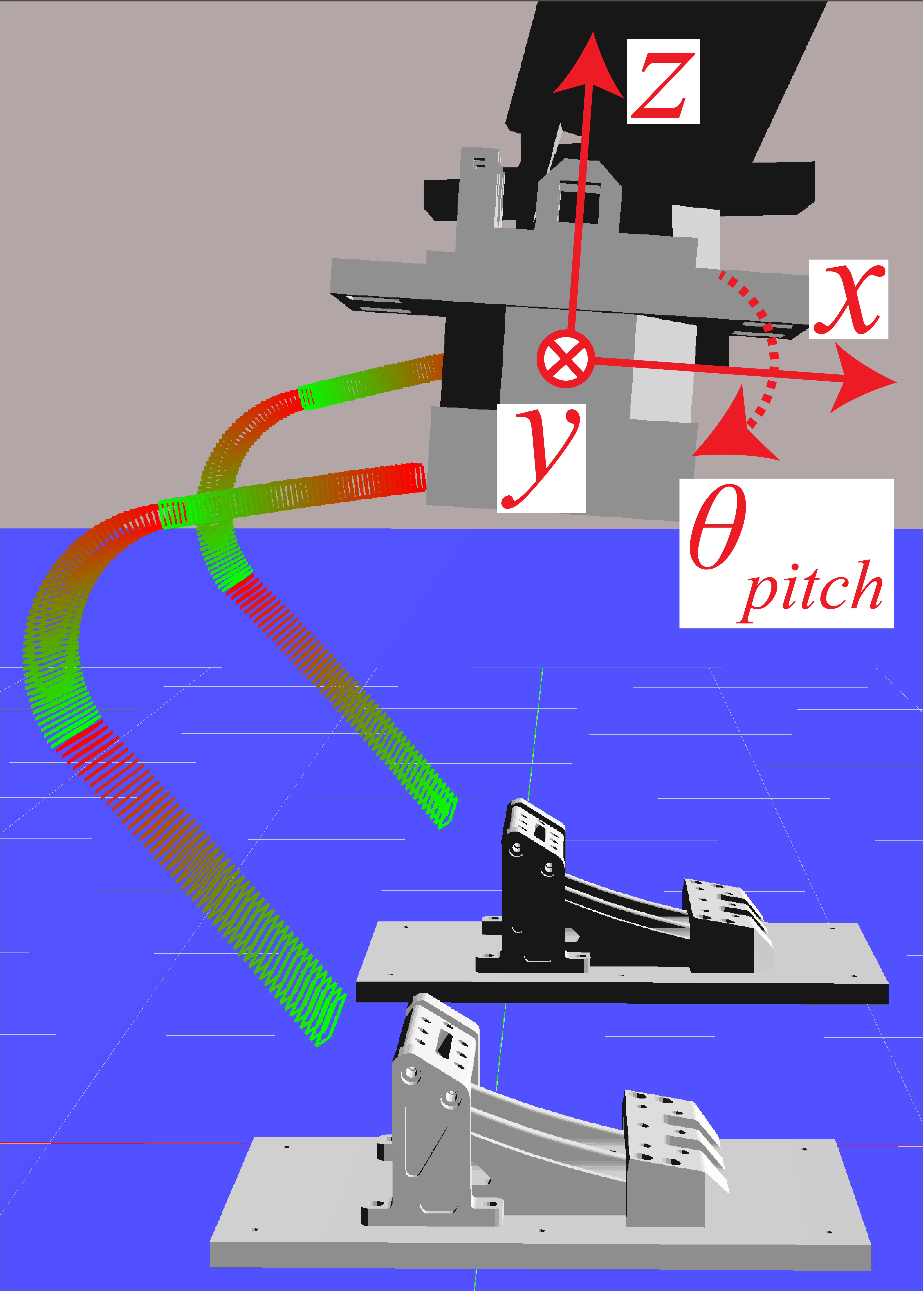}
     \label{fig:foot_coord}
     }\hspace{-3mm}
    \subfigure[]{
    \centering
	 \includegraphics[width=0.73\hsize]{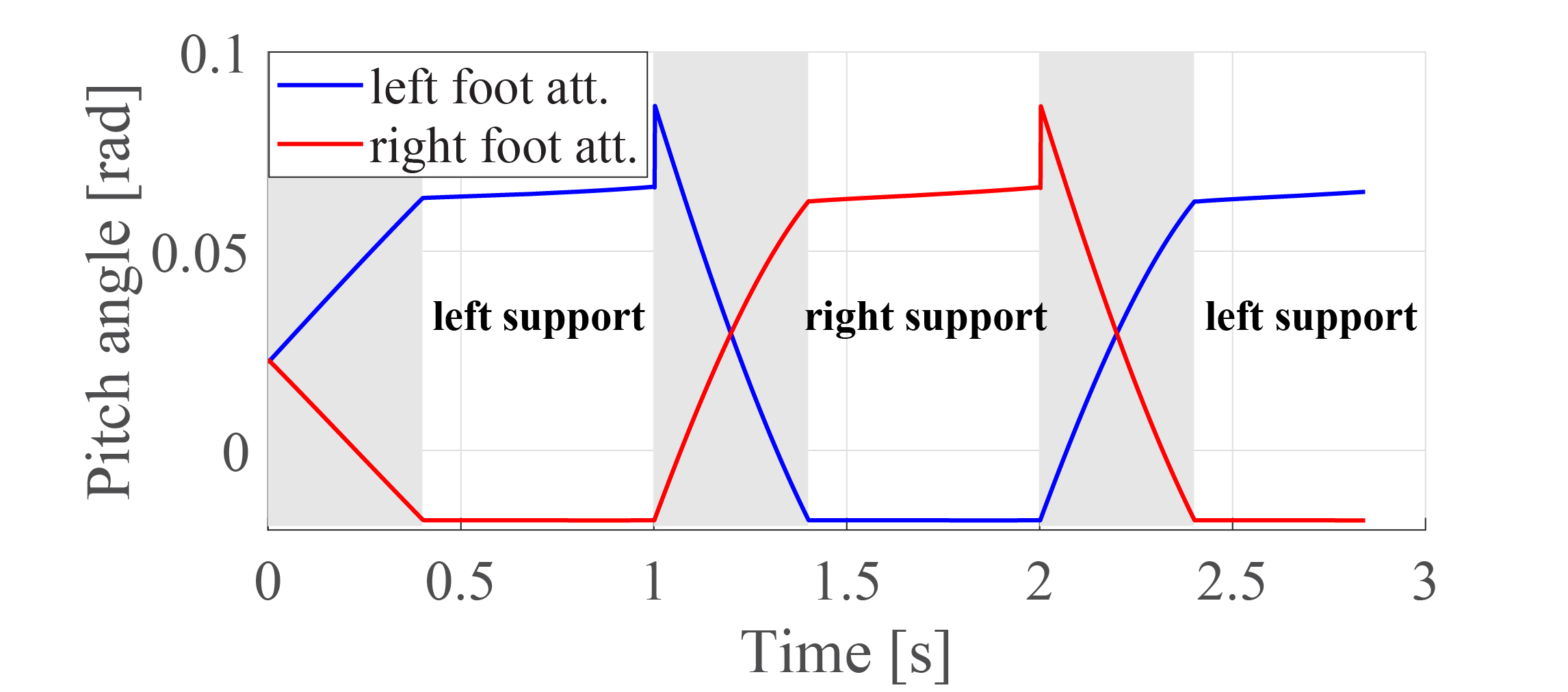}
     \label{fig:foot_att}
     }
	 \caption{(a) Pitch rotation angle plotted to represent the connecting link attitude. (b) Attitude of the connecting link between rigid and soft parts. A larger angle represents larger PCS forward bending deformation. The step duration is 1 s and the double-support period of each step marked by gray regions, is 0.4 s. In this walking motion, the robot first steps forward with its right leg. In each double-support period, the whole-body COG shifts from one leg to the other. 
     }
	\end{center}
\end{figure}
We calculated a biped walking motion with a total of 9 steps for the simulation presented in Section \ref{sect:hybrid_walk}.
The duration was 1.0 per step, including the double-support state with 0.4 s.
The COG height, step length, step duration, and the swing foot maximum height were 1.13 m, 0.05 m, 1 s, and 0.06 m, respectively. 
\figref{fig:foot_att} shows the attitude of the connecting link of each leg in the first 3 steps of walking motion; the blue and red lines indicate these attitudes of the left and right legs, respectively.
The gray area indicates the double-support state.
When the prosthesis is not subjected to external force, the angle of the connecting link was approximately +0.02 rad. 
Because the foot end of the prosthesis is always set at a zero angle, a larger deflection angle of the connecting link indicates a greater deformation in the soft part. 
In each double-support period (gray region), the deformation of the prosthesis shifts with the COG, moving from one leg to the other. In each single-support period, the deformation of the swing leg was at its initial value (+0.02 rad), and the deformation of the support leg remained at the maximum level (of approximately 0.06 rad) due to the gravity. 
We also observed a sudden change of the attitude in transition from a single- to double-support states. 
At 1 s, for example, the left foot attitude suddenly increases to +0.08 rad and then gradually decreases to zero.
We consider that this is because a large ground reaction force is generated on the left foot at the moment of the transition.
The left foot was the supporting foot but suddenly changes to the kicking foot at the moment of the transition.
As the gait planner \cite{sugihara20213d} generates a larger COG acceleration in the forward direction, a large ground reaction force is generated by kicking the ground by the left foot, which results in a sudden deformation of the prosthesis in the left foot.
As shown in Section \ref{sect:simulation}, the proposed control method can stabilize a walking motion despite such a sudden change of the prothesis.

%% file: hdcc.tex
\section{Control of Hybrid-link Humanoid based on Compliance Optimization \label{sect:compliance}}

\subsection{Feasible Solution Space in QP-based Method for Contact Force Distribution\label{sect:qp_based_method_discussion}}
The outline of the QP-based control for a humanoid robot is summarized as follows \cite{Sugihara2020AR}:
\begin{enumerate}
    \item {\it Task-space control} calculates the desired acceleration $\ddot{\bm{p}}_\ast$ assuming a feedback control in the task-space;
    \item {\it Inverse dynamics} simultaneously calculates the necessary joint actuation force and contact forces.
\end{enumerate}
Typically, a task-space feedback control is expressed as
\begin{align}
 \label{eq:task_space_compliance}
 \bm{f} =
 \bm{f}_{\rm ref} +
 \bm{K} \varDelta \bm{p}
 + \bm{D} \varDelta \bm{v}
\end{align}
where, $\bm f$, $\bm f_{\rm ref} \in \mathbb R^M$ are the task-space force and its reference value, respectively; 
$\bm{K}$ and $\bm{D} \in \mathbb{R}^{M \times M}$ are the task-space stiffness and viscosity matrices, respectively; and $\varDelta \bm p$ and $\varDelta \bm v$ denote the task-space position and velocity errors, respectively.

The QP method is often used in the second step to calculate the optimal distribution between joint actuation force and contact forces.
A straightforward implementation is represented as
\begin{align}
\label{eq:qp_for_optimal_contact_force_distribution}
    &\min_{\bm{\tau}_R, \bm{f}_C} \frac{1}{2}
    \| {\tt Inv}(\bm{q},\dot{\bm{q}},\ddot{\bm{q}}) 
    - \bm{\tau} - \mathcal{J}_C^T \bm{f}_C \|^2
\\
  &\mathop{\rm s.t.} \quad \mathcal{C}_1 \bm{\tau}_R \le \bm{d}_1, 
    \quad
    \mathcal{C}_2 \bm{f}_C \le \bm{d}_2. 
\end{align}
where,
\begin{itemize}
    \item ${\tt Inv}(\bm{q},\dot{\bm{q}},\ddot{\bm{q}})$ is an inverse dynamics computation of the left-hand side of the floating-base dynamics \eqref{eq:floating_base_hybrid_system}. Among these, $\bm q$ and $\dot{\bm q}$ are current state, and $\ddot{\bm{q}}$ is calculated from required task acceleration of the first step. 
    \item Generalized force $\bm{\tau}$ is defined as
\begin{align}
    \bm{\tau} := 
    \begin{bmatrix}
        \bm{0}^T & \bm{\tau}_R^T & \bm{\tau}_S^T
    \end{bmatrix}^T;
\end{align}
    \item and $(\mathcal{C}_1, \bm{d}_1)$ and $(\mathcal{C}_2, \bm{d}_2)$ are sets of the coefficient matrix and vector representing constraints on joint actuation and contact forces, respectively.
\end{itemize}
In the case of a rigid-body multi-link system, there is usually sufficient space for the feasible solution of $\bm{\tau}_R$ and $\bm{f}_C$ with appropriate constraints.
In a hybrid-link system, however, the feasible solution space is very limited due to the following reasons.

Substituting \eqref{eq:passive_elasticity} into \eqref{eq:floating_base_hybrid_system} and extracting rows corresponding to $\bm{\tau}_S$, we obtain
\begin{align}
    \label{eq:hybrid_link_contact_force_constraint}
    \mathcal{J}_{C,S}^T \bm{f}_C 
    =
    \begin{bmatrix}
        \bm{M}_{0S}^T & \bm{M}_{RS}^T & \bm{M}_{S}
    \end{bmatrix}
    \ddot{\bm{q}}
    + \bm{\beta}_S
    - \bm{K}_S \varDelta \bm{\theta}_S + \bm{D}_S \dot{\bm{\theta}}_S
\end{align}
where $\mathcal{J}_{C,S}$ is the sub-matrix in $\mathcal{J}_{C}$ corresponding to the PCS model.
Given $\bm{q},\dot{\bm{q}},\ddot{\bm{q}}$ in a hybrid-link system, the value of the right-hand side of this equation is automatically determined by the passive viscoelasticity.
Then, the contact force $\bm{f}_C$ needs to satisfy \eqref{eq:hybrid_link_contact_force_constraint}.
In the example in Section \ref{sect:simulation}, the size of $\mathcal{J}_C^T \in \mathbb R^{n_s N_S \times 3N_C}$ is obtained as follows:
\begin{itemize}
    \item In the single-support, the size was $18 \times 12$ with $n_S = 3$, $N_S = 6$ (the supporting prosthesis legs were modeled using a 6-segment PCS model), and $N_C = 4$ because the single foot sole was modeled with a rectangle shape. 
    \item In the double-support state, the size was $36 \times 32$ with $n_S = 3$, $N_S = 12$, and $N_C = 8$.
\end{itemize}
In each case, $\mathcal{J}_C^T$ was vertically-long, implying that, in \eqref{eq:hybrid_link_contact_force_constraint}, the number of equations is more than that of variables.
Therefore, \eqref{eq:hybrid_link_contact_force_constraint} does not have a solution.
As a result, it is expected that the feasible solution space for \eqref{eq:qp_for_optimal_contact_force_distribution} becomes limited after \eqref{eq:passive_elasticity} is satisfied.

\subsection{Motion Control as Compliance Optimization \label{sect:scc_single}}
We considered passive and active compliance in the hybrid-link system, This is expressed as follows:
\begin{align}
    \label{eq:joint_pd_ctrl}
    \bm{\tau} &= \bm{\tau}_{\rm ref} + \bm{K}_{\theta} \varDelta \bm{\theta} + \bm{D}_{\theta} \varDelta \dot{\bm{\theta}}, 
    \\
    \bm{\tau}_{\rm ref}
    &=
    \begin{bmatrix}
        \bm{\tau}_{R, {\rm ref}} \\ \bm{0}
    \end{bmatrix},
    \bm{K}_{\theta}
    =
    \begin{bmatrix}
        \bm{K}_R & \bm{K}_{RS} \\
        \bm{O} & \bm{K}_S
    \end{bmatrix},
    \bm{D}_{\theta}
    =
    \begin{bmatrix}
        \bm{D}_R & \bm{D}_{RS} \\
        \bm{O} & \bm{D}_S
    \end{bmatrix}
\end{align}
where 
$\bm{K}_{\theta}$ and $\bm{D}_{\theta} \in \mathbb{R}^{N \times N}$ 
are {\it hybrid} stiffness and viscosity matrices, respectively.
Because the formulation on viscosity is similar to that of stiffness, we focused on stiffness for the formulations hereafter.

The hybrid stiffness matrix $\bm{K}_{\theta}$ contains the active passive blocks $\bm{K}_R$ and $\bm{K}_S$, and also an interference $\bm{K}_{RS}$.
This is a feedback term by another active stiffness matrix that determines the active control input $\bm{\tau}_R$ based on $\varDelta \bm{\theta}_S$.
Combining those active stiffness matrices, we define \begin{align}
    \label{eq:active_stiffness}
\widehat{\bm K}_R := \begin{bmatrix}
     \bm K_R & \bm K_{RS} 
 \end{bmatrix}
\end{align}
for subsequent formulations.
Note that lower-left non-diagonal block is a zero matrix because there is no feedback term that determines $\bm{\tau}_S$ based on $\bm{\theta}_S$.

Compliance matrix is defined as the inverse of stiffness: $\bm{C}_{\ast} := \bm{K}_{\ast}^{-1}$.
In particular, hybrid compliance is calculated as
\begin{align}
    \label{eq:scc_decoupling}
 \bm C_{\theta} = \begin{bmatrix}
     \bm C_R & \bm C_{RS} \\ \bm O & \bm C_S
 \end{bmatrix} = \begin{bmatrix}
     \bm K_R^{-1} & -\bm K_R^{-1} \bm K_{RS} \bm K_S^{-1} \\ \bm O & \bm K_S^{-1}
 \end{bmatrix} .
\end{align}
In a manner similar to \eqref{eq:active_stiffness}, we define 
\begin{align}
    \widehat{\bm C}_R
    :=
    \begin{bmatrix}
        \bm C_R & \bm C_{RS} 
    \end{bmatrix}    
\end{align}
Assuming the single-support state, the relationship between the task-space and hybrid compliance is obtained from the virtual work principle as follows \cite{salisbury1980active}:
\begin{align}
    \label{eq:compliance_relationship}
    \bm{J} \bm{C}_{\theta} \bm{J}^T 
    =
    \bm{C}
\end{align}
where $\bm{J} \in \mathbb R^{M\times N}$ is the Jacobian matrix; $\bm C$ is task-space compliance matrix.
We can rewrite this equation as
\begin{align}
\label{eq:separa}
 \bm{J}_R 
 \widehat{\bm C}_R
 \begin{bmatrix}
     \bm J_R & \bm J_S
 \end{bmatrix}^T
    +
    \bm{J}_S \bm{C}_S \bm{J}_S^T
 =
 \bm C
\end{align}
using block elements $\bm{J}_R$ and $\bm{J}_S$ in $\bm J = [ \bm{J}_R \ \bm{J}_S ]$ that correspond to active- and passive-parts, respectively.

The control problem is to calculate an appropriate $\widehat{\bm C}_R$ when $\bm{C}$ and $\bm{C}_S$ are given.
\eqref{eq:separa} can be rewritten as
\begin{align}
\label{eq:static}
 \bm J_R  \widehat{\bm C}_R \bm J^T
 =
     \widehat{\bm C} :=
 \bm C - \bm J_S \bm C_S \bm J_S^T
\end{align}
When there are redundant DOFs, the solution of $ \widehat{\bm C}_R$ in this equation is not uniquely determined.
Therefore, we can calculate a solution using the following optimization:
\begin{align}
    \label{eq:compliance_optimization_single_support}
    \min_{\widehat{\bm C}_R} \frac{1}{2} \| \widehat{\bm{C}}_{R,\rm{ref}} - \widehat{\bm C}_R \|^2
    \quad \mbox{s.t. \eqref{eq:static}} .
\end{align}
where $\| \ast \|^2$ denotes the Frobenius norm for a matrix $\ast$. 
The closed-form solution of this optimization problem is given by
\begin{align}
\label{eq:static_single_sol}
 \widehat{\bm C}_R = \bm J_R^{\sharp} \widehat{\bm C} \bm J^{T\sharp} + \widehat{\bm C}_{R,\rm ref} - \bm J_R^{\sharp}\bm J_R \widehat{\bm C}_{R, \rm ref} \bm J^{T}\bm J^{T\sharp}
\end{align}
\begin{figure}[t]
	\begin{center}
    \subfigure[]{
	 \includegraphics[width=0.25\hsize]{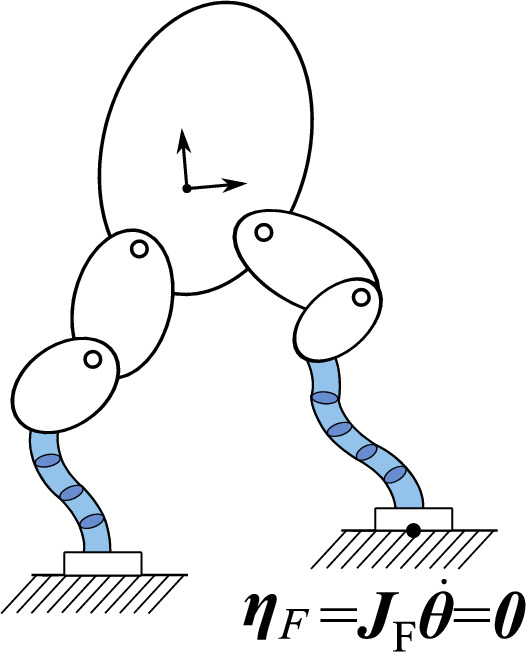}
    \label{fig:hybrid_double}}
  \hspace{10mm}
    \subfigure[]{
    \centering
    \includegraphics[width=0.3\hsize]{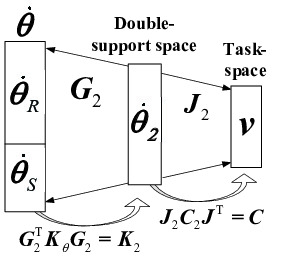}
    \label{fig:double_space}
    }
	 \caption{(a) Illustration of the double-support case. The left foot is considered as a fixed base, and the right foot has an equation constraint on its velocity. (b) Relationship between the different spaces of a hybrid system comprising a closed kinematic chain. The rectangles in the figure represent the velocity spaces. The solid straight arrows show the transformation direction between different velocities. The hollow curved arrows represent the transformation relationship between different space compliance. }
	\end{center}
\end{figure}
\subsection{Compliance Optimization in Double-support State}
In the case of the double-support state, we can assume that a swing foot contacts with the ground, as shown in  \figref{fig:hybrid_double}.
This contact constraint can be represented as
\begin{align}
    \label{eq:double_support_cnstr}
    \bm{\eta}_F = \bm{J}_F \dot{\bm{\theta}} = \bm{0}
\end{align}
where $\bm{\eta}_F \in \mathbb{R}^6$ is the spatial velocity vector of the contacting foot, and $\bm{J}_F$ is its Jacobian matrix.
The dimension of the joint-space can be reduced using this constraint.
$\bm{J}_F$ and $\dot{\bm \theta}$ can be divided as 
\begin{align}
    \bm{J}_F 
    =
    \begin{bmatrix}
        \bm{J}_{F1} & \bm{J}_{F2}
    \end{bmatrix},
    \
    \dot{\bm{\theta}} =
    \begin{bmatrix}
        \dot{\bm{\theta}}_1 \\ \dot{\bm{\theta}}_2
    \end{bmatrix}
\end{align}
where, $\bm{J}_{F1}$ is a square matrix combining six independent column vectors selected from $\bm{J}_F$, and $\bm{J}_{F2}$ is a matrix combining the rest of column vectors.
And, $\dot{\bm{\theta}}_1 \in \mathbb{R}^6$ and $\dot{\bm{\theta}}_2 \in \mathbb{R}^{N-6}$ are elements corresponding to $\bm{J}_{F1}$ and $\bm{J}_{F2}$, respectively.
Because $\bm{J}_{F1}$ has an inverse matrix, we can obtain
\begin{align}
    \label{eq:double_support_mapping}
    \dot{\bm{\theta}} = \bm{G}_2 \dot{\bm{\theta}}_2,
    \
    \bm{G}_2 := 
    \begin{bmatrix}
        -\bm{J}_{F1}^{-1}\bm{J}_{F2} \\
        \bm{E}
    \end{bmatrix} .
\end{align}
In this study, we refer to the space of $\dot{\bm{\theta}}_2$ as the {\it double-support space}.
We can obtain the Jacobian matrix between the task-space velocity $\bm{v}$ and $\dot{\bm{\theta}}_2$ by
\begin{align}
    \label{eq:double_support_task_mapping}
 \bm{v} = \bm{J}_2 \dot{\bm{\theta}_2}, \quad \bm J_2 := \bm J \bm G_2.
\end{align}

\figref{fig:double_space} shows the two mappings performed using \eqref{eq:double_support_mapping} and \eqref{eq:double_support_task_mapping}.
The relationships of the stiffness or compliance among these three spaces are summarized as follows \cite{Yamamoto2017ICRA}:
\begin{itemize}
    \item The mapping of the stiffness between the joint- and double-support-space is given by
\begin{align}
 \label{eq:double_compliance_condition02}
 \bm{G}_2^T \bm{K}_{\theta} \bm{G}_2 = \bm{K}_2 
\end{align}
    \item The mapping of the compliance between the double-support and task-spaces is given by
\begin{align}
    \label{eq:double_compliance_condition00}
    \bm{J}_2 \bm{C}_2 \bm{J}_2^T = \bm{C} .
\end{align}
\end{itemize}
We can first solve \eqref{eq:double_compliance_condition00} followed by \eqref{eq:double_compliance_condition02}.

\subsubsection{Solving \eqref{eq:double_compliance_condition00}}
Although we can solve \eqref{eq:double_compliance_condition00} in a manner similar to \eqref{eq:compliance_optimization_single_support}, we consider another optimization using the inertia matrix as weighting matrix. 
\begin{align}
\label{eq:closed_compliance}
    \min_{\bm{C}_2} \ \frac{1}{2} \| \bm{M}_2^{\frac{1}{2}} (\bm C_{2,\rm ref} - \bm{C}_2) \bm{M}_2^{\frac{1}{2}} \|^2 &
    \quad \mbox{s.t. \eqref{eq:double_compliance_condition00}}
    \\
    \bm{M}_2 := \bm{G}_2^T\bm{M} \bm{G}_2 &
\end{align}
where $\bm{M}_2$ is the inertial matrix in the double-support space.
If we set $\bm C_{2,\rm ref} = \frac{1}{y} \bm{M}_2^{-1}$ for arbitrary $y > 0$, the solution is obtained as
\begin{align}
\label{eq:osc_equivalent_stiff}
 \bm{K}_{2}
 =
 \bm{J}_2^T  \bm{K}\bm{J}_2
 +
 (\bm{E} - \bm{J}_2^T \bm{J}_2^{T\dagger}) \bm K_{2,ref},
 \\
 \bm{J}_2^{\dagger} := \bm{M}_2^{-1} \bm{J}_2^T (\bm{J}_2 \bm{M}_2^{-1} \bm{J}_2^T)^{-1},
\end{align}
where $\bm{J}_2^{\dagger}$ is the dynamically-consistent inverse \cite{Khatib1987} of $\bm{J}_2$.
See \ref{apdx:hybrid_dcc_double} for a detailed derivation of this solution.

If we consider an elastic force $\bm{\tau}_2 = \bm{K}_2 \varDelta \bm{\theta}_2$, it becomes equivalent to the operational space control.
\begin{align}
    \bm{\tau}_2 &= \bm{J}_2^T \bm{f}_2 + (\bm{E} - \bm{J}_2^T \bm{J}_2^{T\dagger}) \bm{y},
    \\
    & \bm{f}_2 := \bm{K}_2 \bm{J}_2 \varDelta \bm{\theta}_2 \simeq \bm{K}_2 \varDelta \bm{p}, \
    \bm{y} := \bm{K}_{2,ref} \varDelta \bm{\theta}_2
\end{align}
where $\bm{f}_2$ and $\bm{y}$ are feedback terms in the task- and null-spaces, respectively.
Therefore, we refer to \eqref{eq:closed_compliance} as {\it dynamically-consistent} compliance optimization (DCC). 
See \ref{apdx:hybrid_dcc_double} for details.

\subsubsection{Solving \eqref{eq:double_compliance_condition02}}
In a manner similar to \secref{sect:scc_single}, we divide $\bm{G}_2$ into two blocks $\bm{G}_{2R} \in \mathbb R^{N_R \times (N_R+3N_S-6)}$ and $\bm{G}_{2S} \in \mathbb R^{3N_S \times (N_R+3N_S-6)}$: $\bm{G}_2 = [ \bm{G}_{2R}^T \ \bm{G}_{2S}^T ]^T$ and rewrite \eqref{eq:double_compliance_condition02} as follows:
\begin{align}
\label{eq33}
    \bm G_{2R}^T &\widehat{\bm K}_R
    \bm{G}_2
    =
    \widehat{\bm K}_2 
    :=
    \bm K_2 - \bm G_{2S}^T \bm K_S \bm G_{2S}.
\end{align}
Then, the optimization problem can be expressed as
\begin{align}
\label{eq:closed_compliance2}
    \min_{\widehat{\bm K}_R} \ \frac{1}{2} \| \widehat{\bm K}_{R,\rm ref} - \widehat{\bm K}_R \|^2
    \quad \mbox{s.t. \eqref{eq33}}
\end{align}
and we can solve it in a manner similar to \eqref{eq:static_single_sol}.
In practice, we set the reference $\widehat{\bm K}_{R,\rm ref}$ as
\begin{align}
    \widehat{\bm K}_{R,\rm ref} = 
    \begin{bmatrix}
    y\bm M_R & \bm O
    \end{bmatrix} .
\end{align}
We set the non-diagonal element to a zero matrix because using a non-zero non-diagonal matrix in the reference stiffness will cause control instability.
Substituting \eqref{eq:osc_equivalent_stiff} into $\bm{K}_2$ in \eqref{eq:closed_compliance2} allows us to solve \eqref{eq:double_compliance_condition00} and \eqref{eq:double_compliance_condition02} in a single procedure.


%% file: simulation.tex
\section{Validations Through Simulations \label{sect:simulation}}

\subsection{Simulation Setting}

The proposed methods were validated using forward dynamics simulations. 
The simulation model is shown in \figref{fig:hybrid_hydra_model}. 
We simulated three scenarios: 1) double-support balancing, 2) vertical hopping, and 3) bipedal walking. 
In the following subsections, we set the viscoelasticity matrices of each PCS segment as $\bm K_S=50\bm E$ and $\bm D_S=0.02\bm E$. 
Note that the viscoelasticity values used here are not derived from actual prosthesis estimated values; rather, they were chosen primarily to achieve more favorable dynamic motion effects in simulation. 
These values are currently determined empirically. 
In future work, these parameters could be further optimized through simulation and serve as a reference for the design of prosthetic legs.
\tbref{tab:subtab11}, \tbref{tab:subtab12}, and \tbref{tab:subtab13} summarizes the parameters used in the simulations. 

\begin{figure}[H]
    \begin{center}
    \centering
        \includegraphics[width=0.9\hsize]{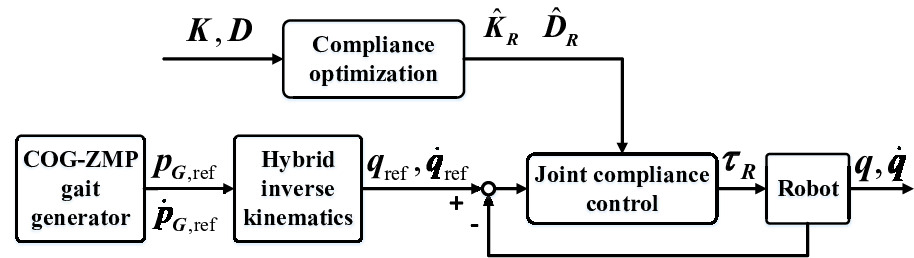}
        \caption{Control framework of a hybrid-link humanoid robot. \label{fig:control_frame}}
    \end{center}
\end{figure}

The general control framework of the proposed method is illustrated in \figref{fig:control_frame}.
As mentioned in \secref{sect:hik_acc}, the COG trajectory is generated by 3D COG-ZMP model \citet{sugihara20213d}. 
In the proposed approach, the matrices $\bm{K}$ and $\bm{D}$ of the COG task are determined based on the capture point feedback control of the COG-ZMP model \cite{sugihara2009standing}. 
The COG task viscoelasticity matrices achieving the feedback are given by $\bm K = - m \omega^2(\mbox{diag}\{\alpha_i\} + \bm{E})$ and $\bm D = - m \omega \ \mbox{diag}\{\alpha_i\}$, where $\omega$ is a constant in the COG-ZMP model, and $\alpha_i < -1 \ (i=x, y, z)$ is the feedback gain parameter. 
The feed-forward torque $\bm{\tau}_{R, \rm ref}$ is set as the gravity compensation torque.

\subsection{Simulations of Standing Balance Control \label{sect:simulation_double}}


\begin{table}[t]
\centering
\caption{Parameter settings in double-support balance control simulations. ($y$ and $z$ are the null-space stiffness and damping values, respectively)}\label{tab:subtab11}
\begin{tabular}{c|c c c}
\toprule
  & COG gain $\alpha_i$ & $y$ & $z$ \\
\hline
PD & / & 100 & 1.0 \\
DCC & 1.0, 1.0, 1000 & 100 & 2$\sqrt{y}$ \\
\bottomrule
\end{tabular}
\end{table}

\begin{table}[t]
\centering
\caption{Parameter settings in Vertical hopping control simulations. ($y$ and $z$ are the null-space stiffness and damping values, respectively)}\label{tab:subtab12}
\begin{tabular}{c|c c c}
\toprule
  & COG gain $\alpha_i$ & $y$ & $z$ \\
\hline
Take-off phase & 1.0, 1.0, 1000 & 300 & 0.2*2$\sqrt{y}$ \\
Flight phase (PD) & / & 500 & 5.0 \\
Touch-down phase & 1.0, 1.0, 1000 & 100 & 2$\sqrt{y}$ \\
\bottomrule
\end{tabular}
\end{table}

\begin{table}[t]
\centering
\caption{Parameter settings in bipedal walking control simulations. ($y$ and $z$ are the null-space stiffness and damping values, respectively)}\label{tab:subtab13}
\begin{tabular}{c|c|c|c|c|c}
\toprule
 & COG & waist & foot & foot & null- \\
 & gain $\alpha_i$ & att. & pos. & att. & space \\
\hline
\hline
 single- & 0.8, 0.8, & k=50, & k=3000, & k=200, & $y$=60 \\
 support & 1000 & d=0.8 & d=1 & d=1 & $z$=10 \\
\hline
 double- & 0.8, 0.8, & \textbackslash & \textbackslash & \textbackslash & $y$=20 \\
 support & 1000 & & & & $z$=8 \\
\bottomrule
\end{tabular}
\end{table}

\begin{figure}[t]
\centering
\subfigure[COG in $x$-axis]{
\includegraphics[width=0.45\hsize]{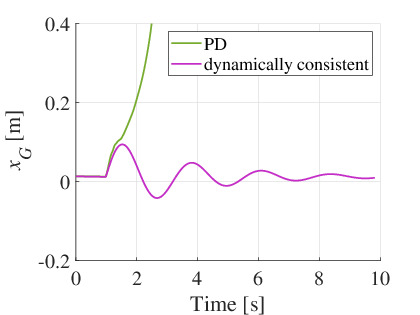}
\label{fig03-3a}
}%
\subfigure[ZMP in $x$-axis]{
\includegraphics[width=0.45\hsize]{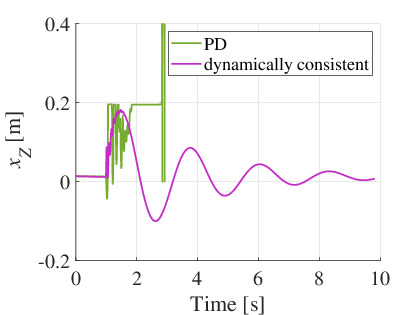}
\label{fig03-3c}
}%
\caption{Simulation results of double support balancing. An external impact (1,000 N $\times$ 0.02 s) was added to the body link in the $x$ direction at 0.2 s. }
\label{fig:double_sim_cog}
\end{figure}

In the double-support standing scenario, we added an external disturbance (1,000 N within 0.02 s) to the body link in the $x$ direction. 
\figref{fig:double_sim_cog} depicts the simulation results, where green and magenta lines indicate the results of the simple PD and DCC optimization, respectively.
In the PD case, the impact caused a large COG and ZMP deviation, and the robot fell down as a result. 
For the DCC case, the standing balance was maintained with a smooth ZMP convergence. 
This result validates that the DCC method optimizes the active compliance considering the variable inertial effect as opposed to constant null-space viscoelasticity in the simple PD case.

\begin{figure}[t]
	\begin{center}
	 \includegraphics[width=0.95\hsize]{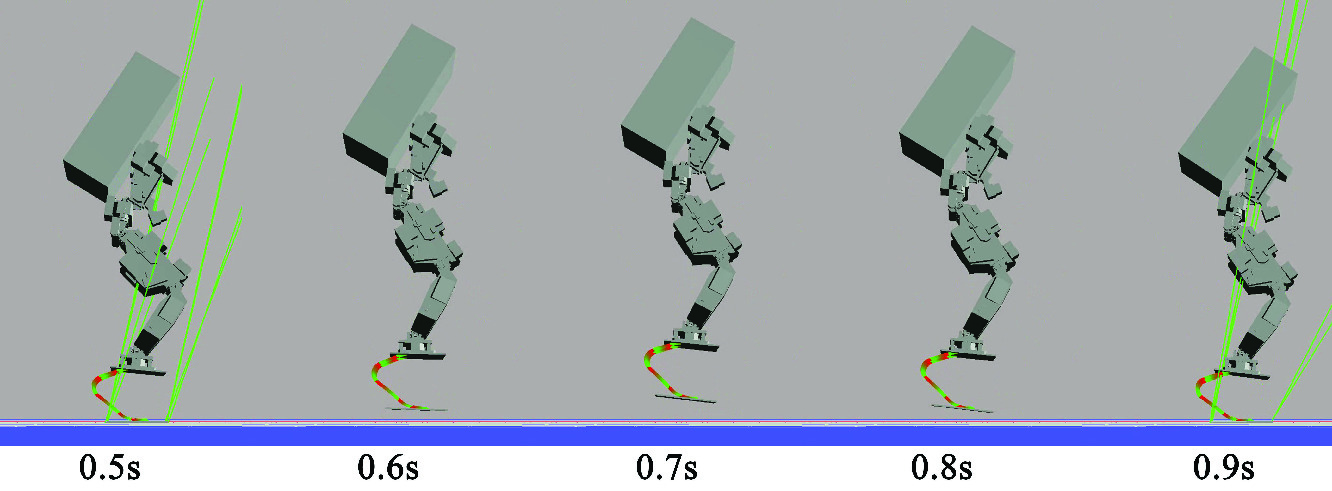}
	 \caption{Snapshots of DCC optimization simulation during vertical hopping and landing (single cycle). The green line shows the ground reaction force on each vertex of the contact plane. \label{fig_snap_jump}}
	\end{center}
\end{figure}

\begin{figure}[t]
\centering
\subfigure[Waist joint torque in pitch]{
\begin{minipage}[t]{0.38\linewidth}
\centering
\includegraphics[width=2.4in]{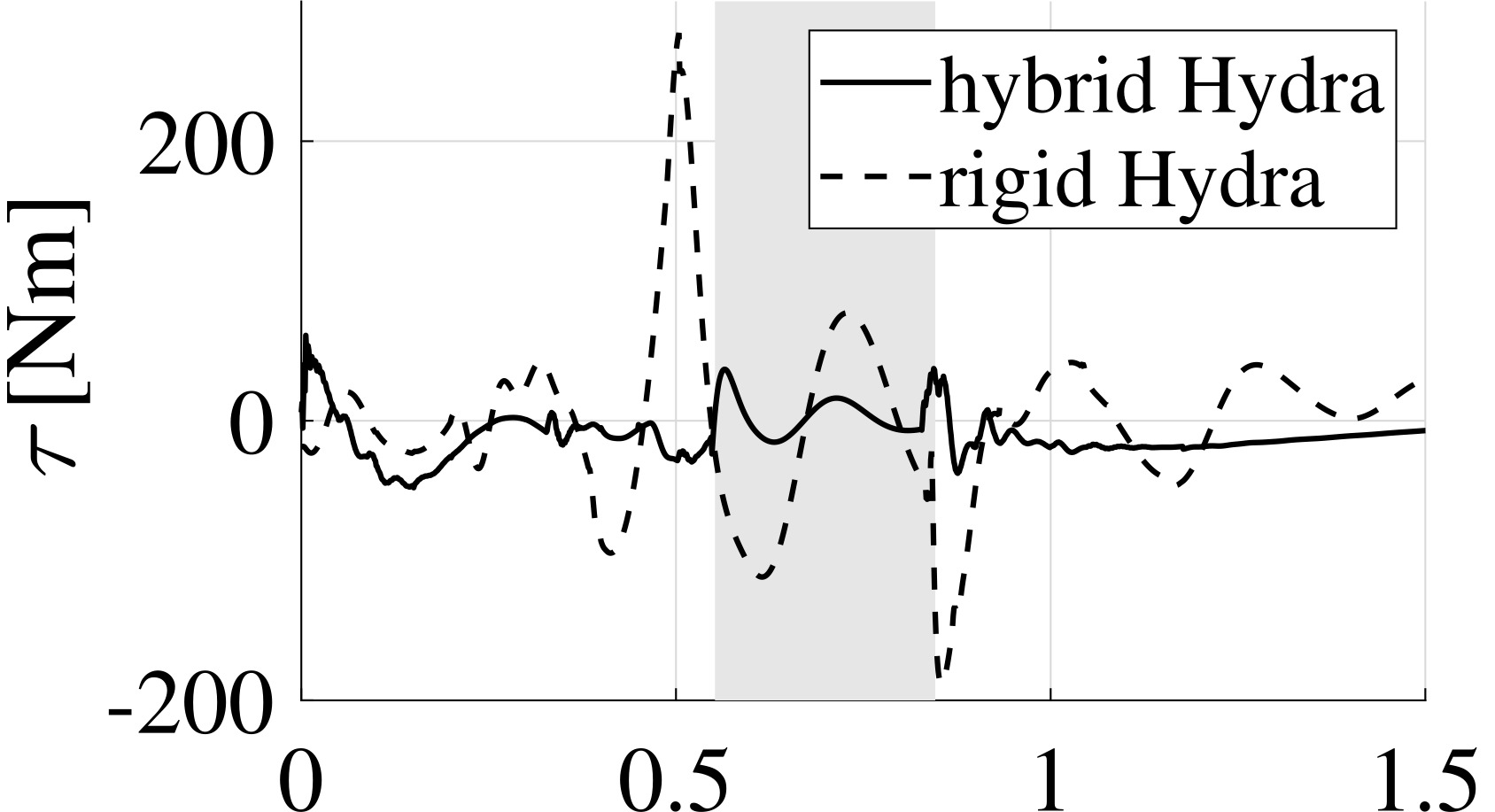}
\end{minipage}%
\label{fig03-4a}
}%
\subfigure[Right hip joint torque in pitch]{
\begin{minipage}[t]{0.67\linewidth}
\centering
\includegraphics[width=2.4in]{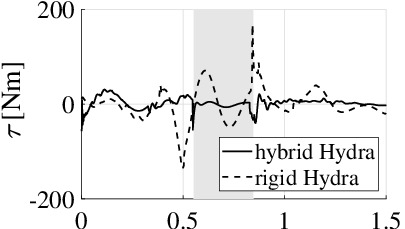}
\end{minipage}%
\label{fig03-4b}
}%

\subfigure[Right knee joint torque]{
\begin{minipage}[t]{0.38\linewidth}
\centering
\includegraphics[width=2.4in]{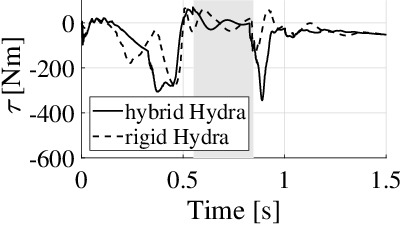}
\end{minipage}%
\label{fig03-4c}
}%
\subfigure[Right ankle joint torque in pitch]{
\begin{minipage}[t]{0.67\linewidth}
\centering
\includegraphics[width=2.4in]{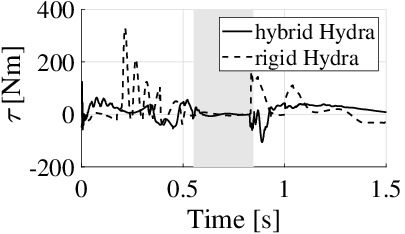}
\end{minipage}%
\label{fig03-4d}
}%
\centering
\caption{Joint torques in the vertical hopping and landing motion. The torque of the left-side joints is similar to that of the right-side joints. The gray region shows the flight phase. }
\label{fig03-4}
\end{figure}

\begin{figure}[t]
	\begin{center}
	 \includegraphics[width=0.7\hsize]{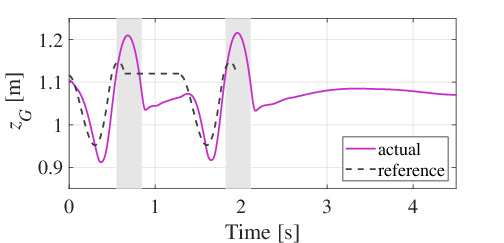}
	 \caption{COG position in the $z$-axis of a twice vertical hopping motion on hybrid Hydra. The dashed black line is the reference trajectory generated by the 3D ZMP model and the magenta line denotes the control result of the simulation. The gray regions show the flight phases in the simulation.  \label{fig_hop_cog}}
	\end{center}
\end{figure}
\subsection{Simulations of Vertical Hopping \label{sect:simulation_jump}}
We demonstrated a continuously twice vertical hopping motion using the proposed DCC control. 
\figref{fig_snap_jump} shows snapshots of the hopping motion during the simulation. 
In the simulation result, the jumping height was 0.11 m at 0.68 and 1.95 s, which is much higher than the reference value (0.03 m at 0.57 and 1.87 s). 
Such a higher jump was obtained because elastic energy was stored in the prosthesis during the take-off phase. 
We simulated a hopping motion using the original Hydra model (without any soft parts) for the same jumping height and duration. 
\figref{fig03-4} shows the waist, hip, knee, and ankle joint torques during the hopping motion, where the black solid and dashed lines indicate the results using the hybrid Hydra and original models, respectively. 
During the take-off phase, it is observed that the peak values of the waist, hip, and ankle torques of the hybrid Hydra were smaller than those of the original model. 
These results provide strong evidence of the effort-reducing and energy-saving performance of flexible elements in dynamic motion. 
\figref{fig_hop_cog} illustrates the COG trajectories in the $z$-axis of this twice hopping motion. 

\subsection{Simulations of Bipedal Walking \label{sect:hybrid_walk}}

\begin{figure}[t]
\centering
\subfigure[COG and ZMP in $x$-axis]{
\centering
\includegraphics[width=0.32\hsize]{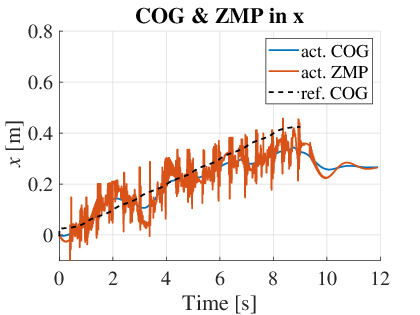}
}\hspace{-3mm}
\subfigure[COG and ZMP in $y$-axis]{
\centering
\includegraphics[width=0.32\hsize]{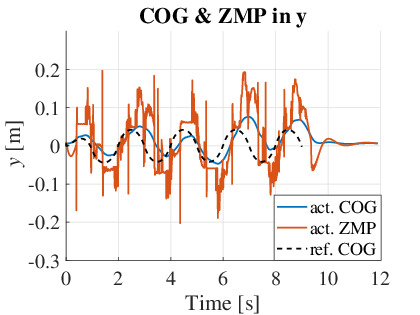}
}\hspace{-3mm}
\subfigure[COG in $z$-axis]{
\centering
\includegraphics[width=0.32\hsize]{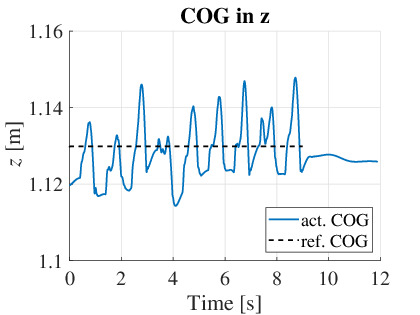}
}%
\centering
\caption{Simulation results of COG (actual and reference) and ZMP of the hybrid walking motion when $\bm K_S = \mbox{diag}\{50\}$. \label{fig:walk_cog}}
\end{figure}

\begin{figure}[t]
    \begin{center}
    \centering
	 \includegraphics[width=0.85\hsize]{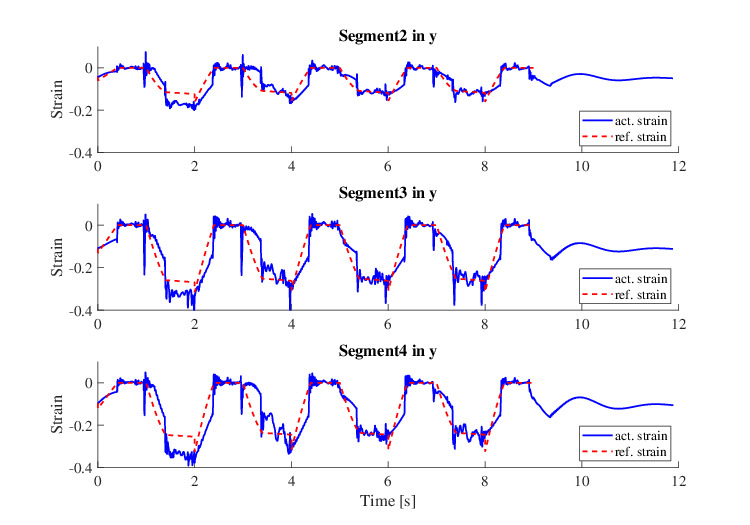}
	 \caption{Actual (forward dynamics simulation) and reference (hybrid IK) PCS strains of right leg during walking motion. \label{fig:pcs_strain}}
	\end{center}
\end{figure}

\begin{figure}[t]
	\begin{center}
	 \includegraphics[width=0.9\hsize]{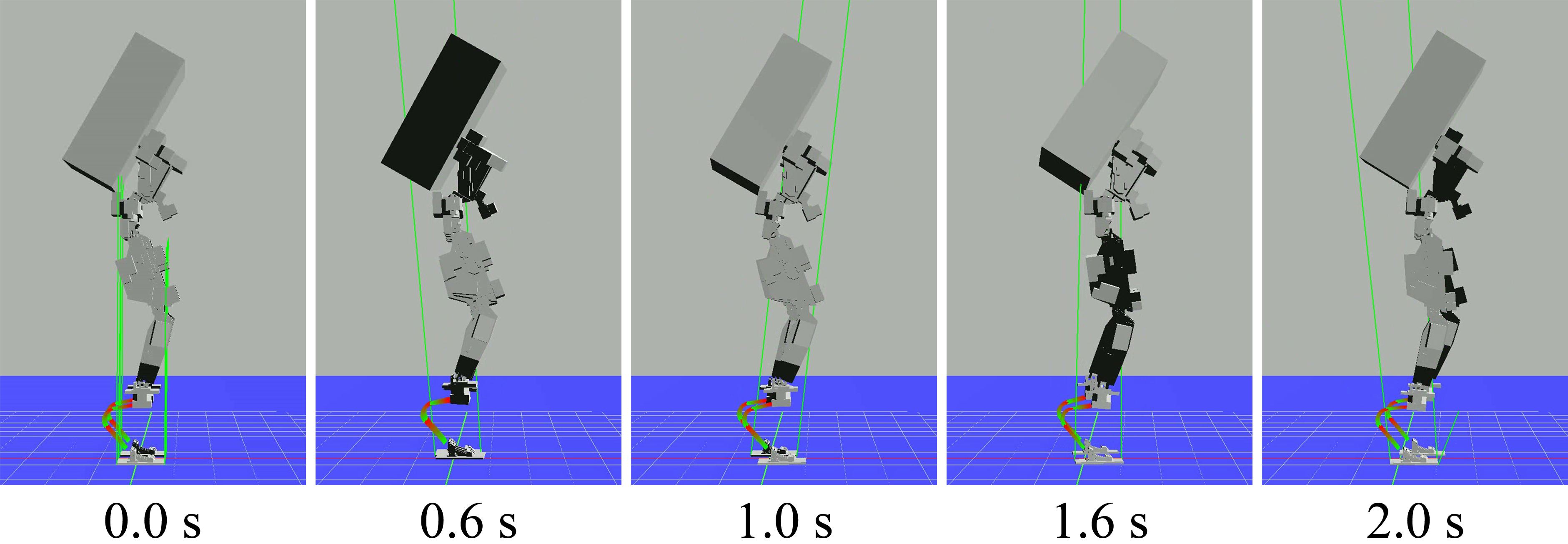}
	 \caption{Snapshots of walking motion control achieved using the proposed method. The first 2 steps are shown. A total of 9 steps were simulated. \label{fig_snap_walk_50}}
	\end{center}
\end{figure}

In this simulation, we applied the walking motion generated in \secref{sec:hik} as a reference, to validate the proposed DCC optimization. 
The COG and ZMP results in the walking simulation are shown in \figref{fig:walk_cog}, and the PCS strain is shown in \figref{fig:pcs_strain}. 
The snapshots of the control simulation are illustrated in \figref{fig_snap_walk_50}. 
Based on the simulation results, we can see that the hybrid Hydra can complete the walking motion. 
Specifically, at the end of the walking motion, the error of COG in $x$-direction is about +0.1 m. 
We calculated the root mean square (RMS) error between the reference PCS strains generated by hybrid IK and the actual strains recorded in the walking simulation. 
The RMS error is 5.6\% during the whole walking motion. 
Based on the COG error and RMS of the strains during the walking motion, the feasibility of the proposed hybrid IK and hybrid compliance optimization method has been validated.

\section{Experimental Validation of Balance Control \label{sect:experiment}}

\subsection{Experimental Setup \label{sec:exp_set}}

\figref{fig:hybrid_hydra} shows an overview of the experimental setup of {\it hybrid Hydra}, which consists of a humanoid robot Hydra \cite{ko2021whole} wearing a leaf-spring type prosthses (1E91 Runner, Ottobock) at each foot. 
Hydra is a life-sized hydro-statically driven humanoid robot, with 14 DoFs, in which all joints are driven by the electro-hydrostatic actuators that are force controllable. 
The total height and weight of the hybrid Hydra are 220 cm and 77 kg, respectively. 

\begin{figure}[t]
    \begin{center}
        \includegraphics[width=0.65\hsize]{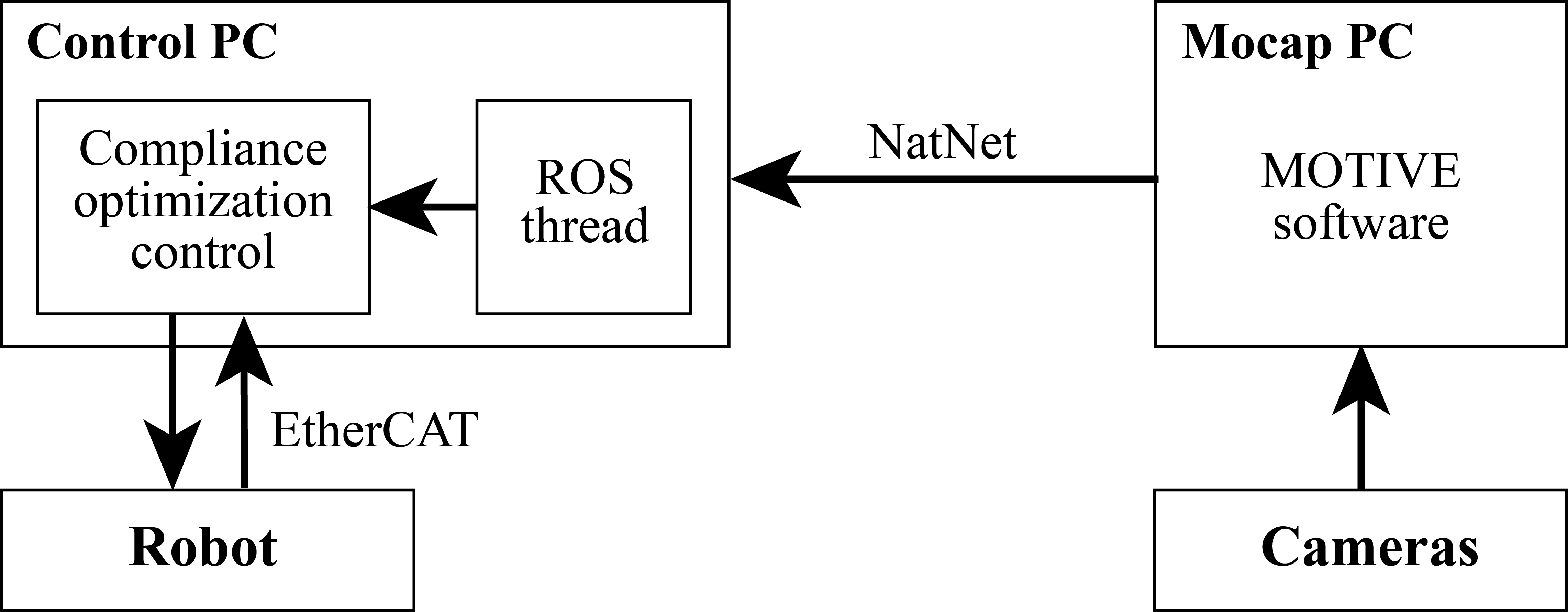}
        \caption{Software architecture of the hybrid Hydra. \label{fig:soft_arch}}
    \end{center}
\end{figure}

We employed a motion capture system for the measurement of the prosthesis deformation, attaching optical markers to the prosthesis.
Although there are a lot of studies on intrinsic sensors \cite{shapiro2013shape,roesthuis2013three}, a motion capture system provides precise measurement of the deformation.
The integrated platform involves two computers: one for robot control (Control PC), and the other for the motion capture system (Mocap PC). 
Six Optitrack$^{\rm TM}$ cameras (Prime$^{\rm X}$ 13W) were connected to the Mocap PC. 
To ensure real-time communication, the two PCs were directly connected through a LAN cable. 
\figref{fig:soft_arch} illustrates the software architecture of the experimental system. 
The total control system runs on the Control PC with a Linux OS (RT-Preempted patched Ubuntu) connecting Hydra through the EtherCAT bus. 
On Mocap PC, the software MOTIVE was installed for basic motion capture data processing, sending the motion capture data to the Control PC through the NatNet protocol. 
The Control PC calculated the IK of the PCS model introduced in \cite{kim2022inverse} using the received data. 
Based on the actual framework, the control frequency of the Control PC was 1 kHz. 
The update frequency of the Mocap PC was approximately 100 Hz, although there was a delay of approximately 4 ms. 

In the experiment, the deformation of each prosthesis leg was divided into three inextensible Kirchhoff-Love segments. 
The optical markers were fixed around the endpoint of each segment (\figref{fig:1e91}). 
The actual calculation time of one-iteration PCS IK was approximately 0.7 ms. 
As a result, the compliance updating frequency is controlled at approximately 300 Hz. 
The control parameters are shown in \tbref{table:exp_gain}. 
In the experiment controller model, the stiffness and viscosity of the PCS segment were $\bm K_S = 15.0 \bm E$ and $\bm D_S = 0.2 \bm E$, respectively. 
Since this controller was directly implemented on the robot, we employed viscoelasticity parameters that closely matched the estimated values for the prosthesis \citet{shimane2024viscoelasticity}. 
In the following experiment results, the measured COG value was calculated by robot kinematics based on measured joint angle and estimated prosthesis deformation. 

\begin{table}[t]
\centering
\caption{Parameter setting of hybrid Hydra for the double-support standing experiment}\label{table:exp_gain}
\begin{tabular}{c c c c}
\toprule
   COG gain $\alpha_i$ & knee task gain & $y$ & $z$ \\
\hline
 0.5, 1.0, 1000 & K=800,\ D=5.0 & 10 & 0.2 \\
\bottomrule
\end{tabular}
\end{table}

\subsection{Whole-body Compliance Experiment}

\begin{figure}[t]
        \centering
        \includegraphics[width=0.9\hsize]{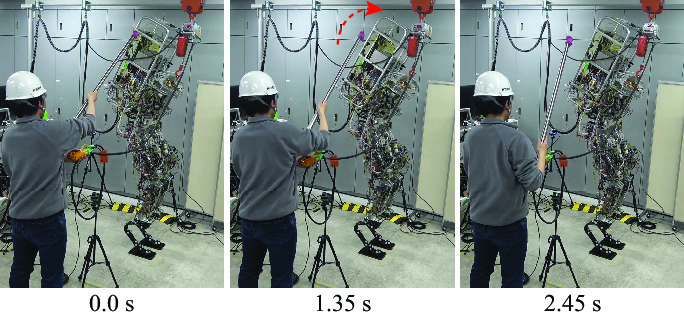}
        \caption{Compliance test with DCC optimization. The red dashed arrows illustrate the motion tendency of the robot. It can be seen that at 1.35 s after being impacted, the robot bends its body along the curved arrow direction to counteract the impact. \label{fig:exp_compliance}}
\end{figure}

In this experiment, the hybrid Hydra was pushed using a stick with a relatively slow and steady force on its body link in $+y$ direction (vertical to the sagittal plane). 
We applied DCC control. 
The snapshots of the experimental video are shown in \figref{fig:exp_compliance}. 
We set the feedback gain of COG in the $x$ direction as 0.5. 
The results showed that when subjected to an external disturbance force, the robot counteracted this disturbance by rotating its body link, which is the null-space of the COG task-space.
Therefore, the DCC optimization demonstrates excellent performance in the null-space compliance. 

\subsection{Push Recovery Experiment}

Based on the previous experiment, we conducted further experiments after changing the parameters of DCC control. 
Specifically, we set the COG gain in the $x$ direction as 5.0. 
In this case, we applied a larger external force to the robot to test its push recovery capabilities. 

\begin{figure}[t]
    \begin{center}
        \includegraphics[width=0.8\hsize]{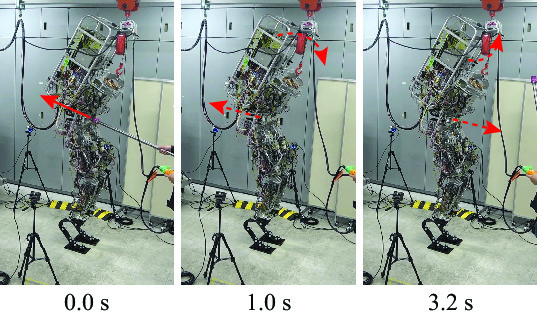}
        \caption{Screenshots of the push recovery experiment applying DCC control. The solid red line shows the external force applied to the robot by a stick. The red dashed lines represent the motion tendency of the robot. After being subjected to an external impact, the robot tilts backward, reaching the maximum COG displacement at approximately 1 s. Simultaneously, during this process, the robot bends its body link to minimize the COG error, which also demonstrates its whole-body compliance. Subsequently, the robot gradually returned to its COG equilibrium position at approximately 3.2 s. During this time, the robot straightened its body and achieved a balanced posture. \label{fig:exp_push_hdcc}}
    \end{center}
\end{figure}

\begin{figure}[t]
    \centering
    \subfigure[]{
        \includegraphics[width=0.5\hsize]{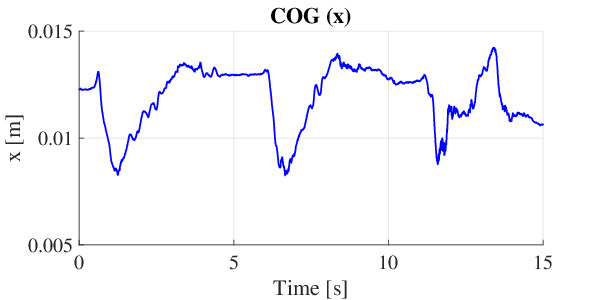}
        \label{fig:hybrid_cog_x}
    }%
    \subfigure[]{
        \includegraphics[width=0.5\hsize]{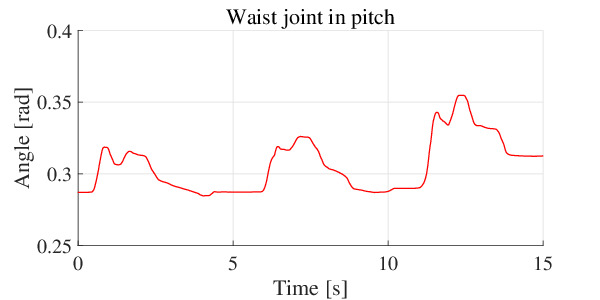}
        \label{fig:hybrid_waist_p}
    }
    \caption{(a) COG trajectories in $x$ directions of the push recovery experiment applying DCC control. In this experiment, three external impacts were applied in the $-x$ direction. From the figures, it can be observed that each time an impact occurs, the COG shifts in the $-x$ direction, reaches its maximum displacement, and then returns to its initial state after approximately 1 to 2 s later. (b) Waist joint angle in $y$ direction (pitch). The measured waist angle increased after impact each time and recovered after several seconds. \label{fig:waist_push_hdcc}}
\end{figure}

Before the experiment applying the proposed controller, we tested the push recovery using a simple PD control. 
The results showed that, after the external impact, the robot fell forward without any body bending. 
While applying the proposed DCC control, as shown in \figref{fig:exp_push_hdcc}, after receiving a large impact, the robot exhibited a pronounced bending of its body and waist links, demonstrating good compliance performance. 
We observed that, upon receiving a backward impact, the robot bends its body forward to compensate for the COG displacement. 
Despite the higher COG gain set in the $x$ direction, the robot did not fall.
The COG trajectory is shown in \figref{fig:hybrid_cog_x}. 
The measured values shifted after receiving the impact and then returned to the initial values when using the proposed control method. 
The angle of the waist joint in pitch is shown in \figref{fig:hybrid_waist_p}, which shows that the actual angle underwent a noticeable increase followed by a decrease. 
This is because the robot bent its body along the pitch direction in response to the impact. 
This result is consistent with the experimental phenomenon shown in \figref{fig:exp_push_hdcc}. 
During the experiment, the COG displacement (in the $x$ direction) was controlled within 0.5 cm. 


%% file: appendix.tex


\section{Hybrid IK in Double-support State\label{apdx:hybrid_ik_double}}
In the case of the double-support state, we can consider an optimization similar to \eqref{eq:ikop2} but with additional constraints on the foot.
\begin{align}
    \min_{\bm{\theta}} \ g(\bm{\theta}) \quad
    \mbox{subject to} \
    \varDelta \bm{p}_F
    = \bm{0}
\end{align}
To discuss the necessary condition of this problem, we define Lagrangian $\mathcal{L}(\bm{\theta})$ as
\begin{align}
    \mathcal{L}(\bm{\theta}) := \bm{g}(\bm{\theta}) + \bm{\lambda}^T \varDelta \bm{p}_F(\bm{\theta}) 
\end{align}
where $\bm{\lambda} \in \mathbb R^6$ is the Lagrange multiplier.
The necessary condition of this optimization is 
\begin{align}
    \label{eq:necessary_condition_lagrange00}
    \left({\pdif{\mathcal{L}(\bm{\theta})}{\bm{\theta}}}\right)^T = \bm{0},
    \ \mbox{and} \
    \left({\pdif{\mathcal{L}(\bm{\theta})}{\bm{\lambda}}}\right)^T = \bm{0}
\end{align}
The first condition in \eqref{eq:necessary_condition_lagrange00} can be rewritten as
\begin{align}
    &\left({\pdif{\mathcal{L}(\bm{\theta})}{\bm{\theta}}}\right)^T 
    =
    \nabla g^T(\bm{\theta}) - \bm{J}_F^T \bm{\lambda}
    \\
    \label{eq:necessary_condition_lagrange01}
    &= -
    \begin{bmatrix}
        \bm{J}_R^T \\ \bm{J}_{S_1}^T \\ \bm{J}_{S_2}^T
    \end{bmatrix}
    \bm{W}_1 \varDelta \bm{p}_G
    -
    \bm J_S^T(\bm \theta) \begin{bmatrix}
        \bm{O} \\ \bm{K}_{S_1} \varDelta \bm{\theta}_{S_1} \\ \bm{K}_{S_2} \varDelta \bm{\theta}_{S_2}
    \end{bmatrix}
    \nonumber
    \\
    & \hspace{6em}+
    \begin{bmatrix}
        \bm{J}_{G,R}^T \\ \bm{J}_{G,S_1}^T \\ \bm{J}_{G,S_2}^T
    \end{bmatrix}
    \bm{g}m -
    \begin{bmatrix}
        \bm{J}_{F,R}^T \\ \bm{J}_{F,S_1}^T \\ \bm{J}_{F,S_2}^T
    \end{bmatrix}
    \bm{\lambda}
    =
    \bm{0}
\end{align}
In this equation, $\bm{\lambda}$ can be regarded as the contact force on the additional foot.
The bottom two rows in \eqref{eq:necessary_condition_lagrange01} represent the equilibrium among the elastic forces by two soft parts, the gravity effect, and the contact force.

\color{black}

\section{Derivation of \eqref{eq:necessary_condition03} and \eqref{eq:necessary_condition04}\label{apdx:necessary_condition_derivation}}

Taking the derivative of \eqref{eq:cog_and_connecting_point},
\begin{align}
\label{eq:deriv_pg}
    \pdif{\bm p_G}{\bm \theta} = \pdif{\bm p_C}{\bm \theta} + \pdif{\bm R_C(\bm \theta_S)}{\bm \theta}{}^{C}\bm{p}_{CG}(\bm{\theta}_R) + \bm R_C(\bm \theta_S) \pdif{{}^{C}\bm{p}_{CG}}{\bm \theta}
\end{align}
where Jacobian matrices can be represented as follows:
\begin{align}
    \bm J_C &:= \pdif{\bm p_C}{\bm \theta} = \begin{bmatrix}
        \bm O & \bm J_{C,S}
    \end{bmatrix}, \\
    \pdif{\bm R_C}{\bm \theta}{}^{C}\bm{p}_{CG} &:= - [(\bm R_C {}^{C}\bm{p}_{CG})\times ]
    \begin{bmatrix}
        \bm O & \bm J_{\omega,S}
    \end{bmatrix}, \\
    \bm J_{C,G} &:= \pdif{{}^{C}\bm{p}_{CG}}{\bm \theta} = \begin{bmatrix}
        \bm J_{CG,R} & \bm O 
    \end{bmatrix}
\end{align}
Finally, we can rewrite $\nabla g^T(\bm{\theta}_\ast) = \bm{0}$ based on \eqref{eq:deriv_pg} as \eqref{eq:necessary_condition03} and \eqref{eq:necessary_condition04}. 

\section{Derivation of DCC Optimization\label{apdx:hybrid_dcc_double}}
\eqref{eq:closed_compliance} is equivalent to
\begin{align}
        \min_{\widetilde{\bm{C}}_2} \ \frac{1}{2} \| y^{-1} \bm{E} - \widetilde{\bm{C}}_2 \|^2 
    \quad \mbox{s.t.} \ 
    \widetilde{\bm{J}}_2 \widetilde{\bm{C}}_2 \widetilde{\bm{J}}_2^T
    \\
    \label{eq:inertia_weighted_mapping}
    \widetilde{\bm{C}}_2 := \bm{M}_2^{\frac{1}{2}} \bm{C}_2 \bm{M}_2^{\frac{1}{2}},
    \quad
    \widetilde{\bm{J}}_2 := \bm{J} \bm{M}_2^{-\frac{1}{2}} .
\end{align}
General solution of this problem is represented by
\begin{align}
    \widetilde{\bm{C}}_2 
    =
    \widetilde{\bm{J}}_2^{\sharp} \bm{C} \widetilde{\bm{J}}_2^{T\sharp}
    +
    (\bm{E} - \widetilde{\bm{J}}_2^{\sharp} \widetilde{\bm{J}}_2) y^{-1},
\end{align}
From the mapping given by \eqref{eq:inertia_weighted_mapping}, this solution can be rewritten as
\begin{align}
    \label{eq:dcc_compliance_solution}
    \bm{C}_2
    =
    \bm{J}_2^{\dagger} \bm{C} \bm{J}_2^{T\dagger}
    +
    (\bm{E} - \bm{J}_2^{\dagger}\bm{J}_2) \bm{M}_2^{-1} y^{-1}
\end{align}
Note that this is the general solution of the dynamically consistent compliance formulation \cite{yamamoto2021humanoid} given by $\bm{J}_2\bm{C}_{2} = \bm{C}\bm{J}_2^{T\dagger}$.

We can check 
\begin{align}
    \bm{K}_2 \bm{C}_2 = \bm{C}_2 \bm{K}_2 = \bm{E}
\end{align}
is satisfied for $\bm{K}_2$ given by \eqref{eq:osc_equivalent_stiff} and $\bm{C}_2$ given by \eqref{eq:dcc_compliance_solution}.
In other words, $\bm{K}_2 = \bm{C}_2^{-1}$ is satisfied.
Therefore, \eqref{eq:osc_equivalent_stiff} is the solution of \eqref{eq:closed_compliance}.
